\documentclass[runningheads]{llncs}

\usepackage{eccv}

\usepackage{eccvabbrv}

\usepackage{graphicx}
\usepackage{booktabs}

\usepackage[accsupp]{axessibility}  % Improves PDF readability for those with disabilities.

\usepackage{url}

\usepackage[utf8]{inputenc} % allow utf-8 input
\usepackage[T1]{fontenc}    % use 8-bit T1 fonts
\usepackage{url}            % simple URL typesetting
\usepackage{booktabs}       % professional-quality tables
\usepackage{amsfonts}       % blackboard math symbols
\usepackage{nicefrac}       % compact symbols for 1/2, etc.
\usepackage{microtype}      % microtypography
\usepackage{xcolor}         % colors
\usepackage{caption} 
\usepackage{colortbl}

\usepackage{amsmath}
\usepackage{colortbl}
\usepackage{algorithm}
\usepackage{algpseudocode}
\usepackage{algorithmicx}
\usepackage{algcompatible}
\usepackage{cancel}
\usepackage{rotating}
\usepackage{placeins}
\usepackage{caption} 

\usepackage{wrapfig}
\usepackage{graphicx}
\usepackage{enumitem}
\usepackage{multirow}
\usepackage{makecell} % preamble에 추가 필요
\usepackage{adjustbox}

\usepackage{subcaption}
\usepackage[table]{xcolor}
\usepackage{tcolorbox}
\tcbuselibrary{theorems}

\usepackage[pagebackref,breaklinks,colorlinks,citecolor=eccvblue]{hyperref}
\usepackage{hyperref}

\usepackage{orcidlink}

\newcommand{\img}{x_{\text{img}}}
\newcommand{\txt}{x_{\text{txt}}}

\begin{document}

% ---------------------------------------------------------------
% TODO REVIEW: Replace with your title
\title{Two Birds, One Projection:\\ Harmonizing Safety and Utility in LVLMs\\via Inference-time Feature Projection} 

% TODO REVIEW: If the paper title is too long for the running head, you can set
% an abbreviated paper title here. If not, comment out.
\titlerunning{TBOP: Safety–Utility Harmonization in LVLMs via IFP}

% TODO FINAL: Replace with your author list. 
% Include the authors' OCRID for the camera-ready version, if at all possible.
% \author{Yewon Han\inst{1}\orcidlink{0000-1111-2222-3333} \and
% Yumin Seol\inst{2}\orcidlink{1111-2222-3333-4444} \and
% Minsoo Jo\inst{1}\orcidlink{2222--3333-4444-5555} \and 
% EunGyung Kong\inst{3}\orcidlink{2222--3333-4444-5555} \and \\
% Taesup Kim\inst{1}\orcidlink{2222--3333-4444-5555}}
\author{Yewon Han\inst{1} \and
Yumin Seol\inst{1}\thanks{Work done during internship. $^\dagger$ Corresponding author.} \and
EunGyung Kong\inst{2} \and 
Minsoo Jo\inst{1} \and 
Taesup Kim\inst{1}$^\dagger$}

% TODO FINAL: Replace with an abbreviated list of authors.
\authorrunning{Y. Han et al.}
% First names are abbreviated in the running head.
% If there are more than two authors, 'et al.' is used.

% TODO FINAL: Replace with your institution list.
\institute{Graduate School of Data Science, Seoul National University
% \and
% Pohang University of Science and Technology 
\and
Mobilint
}
\maketitle

\begin{abstract}
Existing jailbreak defense frameworks for Large Vision-Language Models (LVLMs) often suffer from a safety–utility trade-off, where strengthening safety inadvertently degrades performance on general visual-grounded reasoning tasks. In this work, we investigate whether safety and utility are inherently antagonistic objectives. We focus on a modality-induced bias direction consistently observed across datasets, which arises from suboptimal coupling between the LLM backbone and visual encoders. We further demonstrate that this direction undermines performance on both tasks. Leveraging this insight, we propose \textbf{\textsc{TBOP}} (\textbf{\underline{T}}wo \textbf{\underline{B}}irds, \textbf{\underline{O}}ne \textbf{\underline{P}}rojection), an efficient inference-time jailbreak defense that projects cross-modal features onto the null space of the identified bias direction to remove the corresponding components. Requiring only a single forward pass, our method effectively breaks the conventional trade-off, simultaneously improving both safety and utility across diverse benchmarks.
\end{abstract}

\section{Introduction}
\label{sec:intro}

Large Vision-Language Models (LVLMs) extend the reasoning capabilities of Large Language Models (LLMs) by integrating visual perception.
However, this expanded input space introduces complex cross-modal interactions during contextualization, which can shift cross-modal features away from the original safety-aligned space established in LLM backbone training. 
Consequently, the model's inherent refusal mechanisms for harmful cross-modal queries are often weakened, significantly increasing susceptibility to jailbreaks.

\begin{wrapfigure}{r}{0.45\linewidth}
    \centering
    \includegraphics[width=\linewidth]{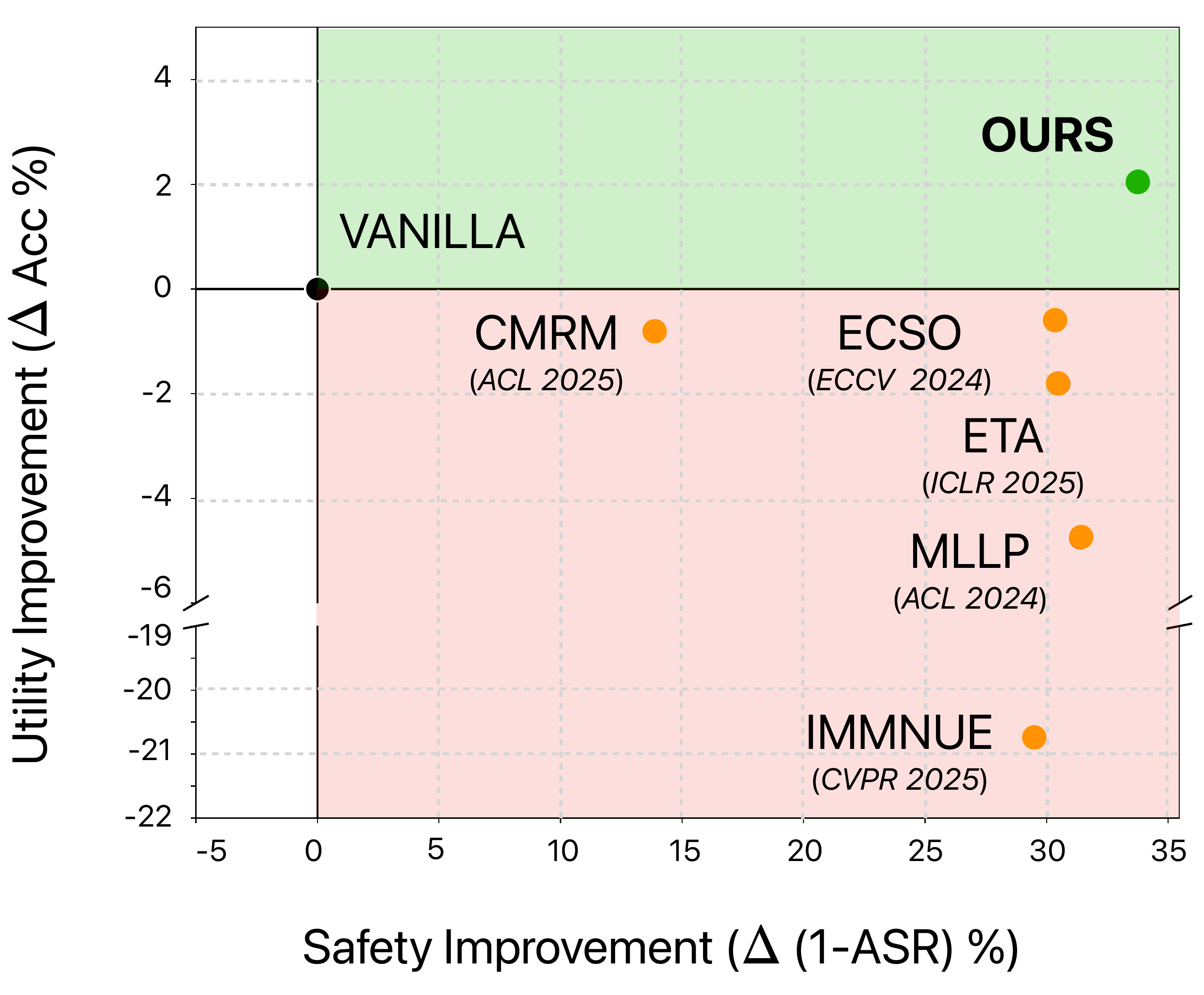}
    \caption{\textbf{Trade-off Between Safety and Utility Observed in LVLM Defense Methods.} By effectively eliminating the shared performance underlying direction, our approach generates safe and useful responses, thereby overcoming the conventional tradeoffs.} 
    \label{fig:f0}
\end{wrapfigure}

Prior research has addressed these safety risks through two primary lenses. Training-based methods such as MLLM-Protector \cite{pi2024mllmprotectorensuringmllmssafety} integrate auxiliary safety modules (e.g., harm detectors) using additional fine-tuning data.
Conversely, inference-time approaches like ECSO \cite{gou2024eyesclosedsafetyon} and ETA \cite{ding2024eta} assess the harmfulness of generated responses and refine them without retraining.
Despite their effectiveness, these approaches predominantly follow a \textit{detect-then-detoxify} paradigm. 
This reliance on post-hoc safety detection inherently triggers a \textit{safety–utility trade-off}.
By treating the symptoms (i.e., harmful outputs) rather than the underlying cause, these methods inevitably become more conservative as defenses strengthen, unnecessarily suppressing benign queries and degrading overall utility performance as observed in \cref{fig:f0}.
Furthermore, this sequential paradigm incurs significant computational overhead and memory costs due to multiple forward passes and auxiliary model dependencies.

While recent generative approaches such as Immune \cite{ghosal2025immune} and CMRM \cite{liu-etal-2025-unraveling-mitigating} attempt to mitigate these risks internally, they still fall short of a holistic solution.
Specifically, Immune lacks a rigorous analysis of the side effects that its internal manipulations have on utility, resulting in substantial performance loss.
CMRM relies on carefully tuned coefficients, making it sensitive to dataset-specific variations thus limiting its effectiveness and generalizability. 
Consequently, these methods still fail to effectively harmonize two objectives: \textbf{safety and utility}.

In this work, we argue that the persistent conflict between safety and utility is not an inherent property of LVLMs, but rather stems from a shared underlying bias caused by the imperfect coupling between the vision encoder and LLM backbone. Unlike prior generative methods that intervene without a unified diagnostic framework, we identify the modality-induced bias as an axis that simultaneously undermines safety and utility performance. 
Our empirical evidence supports this, highlighting that the two objectives are not antagonistic; rather, they are jointly compromised by this common structural root cause.

Motivated by this insight, we propose \textsc{TBOP}, an efficient inference-time defense strategy for LVLMs that can generate \textbf{safe and useful} responses.
\textsc{TBOP} removes the bias-related components from cross-modal features by projecting them onto the null space of the identified bias direction. 
Targeting the principled direction rather than the overall bias allows our method to remain effective across diverse datasets, unaffected by inconsistent bias intensity. 
Extensive experiments demonstrate that our simple projection-based approach simultaneously improves performance across diverse safety and utility benchmarks, effectively breaking the conventional tradeoff and establishing a more robust foundation for secure LVLM deployment. Since such biases naturally emerge when combining modules that process different modalities, and are costly to mitigate with additional training, \textsc{TBOP} provides an efficient inference-time remedy that reduces this overhead.

Our main contributions can be summarized as follows:
\begin{itemize}
    \item We identify that modality-induced bias consistently affects both safety and utility performance, serving as a shared performance-undermining direction.
    \item We propose an effective defense strategy for LVLMs that can achieve significant improvements across a wide range of safety and utility tasks.
    \item \textsc{TBOP} is designed as a computationally efficient inference-time defense strategy that requires only a single forward pass, enabling seamless integration.
\end{itemize}

\section{Related Work}
\label{sec:related}

\paragraph{\textbf{Large Vision Language Models.}}
Recent advances in large vision language models (LVLMs) have demonstrated that pairing visual encoders with pretrained text-only LLMs, typically via a projector that bridges the visual feature space and the LLM’s token-embedding space.

Representative systems include LLaVA\cite{liu2023visual}, InstructBLIP\cite{dai2023instructblip}, Qwen2-VL\cite{wang2024qwen2}, and InternVL\cite{chen2024internvl}, which adopt this general design while differing in the specific alignment modules or training strategies for integrating visual representations with language models.
Such modular coupling often induces representational misalignment, arising from the components being optimized under different objectives and training stages\cite{liu2025reducing, jing2025comprehensive, zhang2024unveiling}, which in turn brings a range of issues, from object hallucination to broader safety vulnerabilities.

\paragraph{\textbf{Jailbreak Attacks and Defenses in LVLMs.}}

LVLMs remain highly vulnerable to jailbreak attacks that exploit cross-modal interactions.
HADES\cite{li2024images} shows that images can conceal and amplify malicious context to evade guardrails, undermining safety alignment. Test-time backdoors like AnyDoor\cite{lu2024test} implant universal triggers via adversarial images, exposing the limits of input filtering.
Recent analyses\cite{liu2024mm, liu2025dream, shayegani2023jailbreak} argue risks compound across modalities\cite{wang2024white, ying2025jailbreak}, demanding defenses that address risk composition rather than single-channel perturbations.

Defenses for LVLM jailbreaks can be broadly grouped into train-based and inference-time approaches. Train-based defenses, including finetuning-based safety alignment\cite{zong2024safety, lu2025adversarial, pi2024mllm}, improve harmlessness but require training cycles and computational overhead. Inference-time defenses, such as prompting strategies\cite{shao2024refusing, wang2024adashield, zhao2024bluesuffix} and alignment methods \cite{ghosal2025immune, ding2024eta} avoid modifying model weights but often suffer from brittleness under cross-modal attacks.

\paragraph{\textbf{Feature Manipulation.}}
Recent work shows that feature-level manipulation can systematically steer LLMs toward safer and more faithful outputs\cite{yu2024robust, park2025steer, wang2024inferaligner}.
ReFAT\cite{yu2024robust} finds a ``refusal'' feature dimension in the residual stream and adversarially trains on its worst-case offset, operationalizing harmful subspaces to improve robustness.

TSV\cite{park2025steer} adds a vector that reshapes representations to split truthful from hallucinatory content without weight updates. 
InferAligner\cite{wang2024inferaligner} uses safety steering vectors from an aligned model to guide a target model.
On the other hand, layer-wise analyses\cite{skean2025layer} further motivate interventions on token-level states rather than only outputs.
Within MLLMs, VTI\cite{liu2025reducing} reduces hallucinations by test-time steering of multimodal features.

Nullu\cite{yang2025nullu} builds a HalluSpace from contrasts between hallucinated and truthful feature embeddings, and suppresses hallucinations via weight editing, yielding an effect akin to feature manipulation. CMRM \cite{wang2025steering} and ShiftDC \cite{zou2025understandingrectifyingsafetyperception} utilize feature manipulation for safety alignment.
Building on these ideas, we propose a defense method that removes bias components in cross-modal features to better align safety while extracting refined representations for enhanced visual grounding.

\section{Consistent Modality-induced Bias as Degradation Axis in LVLMs.}

\subsection{Preliminaries}
\label{subsec:preliminaries}
\paragraph{\textbf{Decomposition of Cross-modal Representations.}}
Following the formulation introduced by Liu et al. \cite{liu-etal-2025-unraveling-mitigating}, the representation of a text–image pair can be decomposed as:
\begin{align}    
    h(\txt,\img) = h^*(\txt,\img) + \alpha\cdot[h(\txt,\img^\prime) - h(\txt)],
    \label{eq:mis}
\end{align}
where the difference term $h(\txt,\img^\prime) - h(\txt)$ defines the representation shift introduced by the visual modality, capturing how the feature changes purely due to the activation of the visual input. The dummy image $\img^\prime$ serves as a blank input to isolate modality activation from visual content, and the ideal component $h^*(\txt,\img)$ represents how the model should encode cross-modal information in the absence of such undesirable distortion.

Because this shift arises from a single underlying mechanism, namely the activation of the visual modality, it induces a consistent pattern across inputs. Consequently, the resulting shifts do not disperse arbitrarily in the feature space but concentrate along a limited set of dominant directions, forming a structured subspace. We refer to this systematic deviation from the ideal representation as \textit{modality-induced bias}, which we argue is a key factor underlying performance degradation in both safety defense and general visual-grounded reasoning.

\paragraph{\textbf{Visual Inputs as Expanded Attack Surfaces.}}
Empirical studies have revealed that visual inputs significantly increase jailbreak vulnerability: HADES \cite{li2024images} reports that even blank images can double attack success rates. Motivated by these findings, defenses such as ECSO \cite{gou2024eyesclosedsafetyon} attempt to mitigate this risk by converting harmful images into text captions and regenerating responses if the initial output is classified as harmful, underscoring the strong link between the visual modality and multimodal vulnerability. Furthermore, CMRM \cite{liu-etal-2025-unraveling-mitigating} analyzes these vulnerabilities in feature level, demonstrating that visual inputs can shift representations away from safety-aligned regions, thus inducing toxic outputs.

\subsection{Observations}
\label{subsec:observations}

\begin{figure*}[t!]
    \centering
        \includegraphics[width=1.0\linewidth]{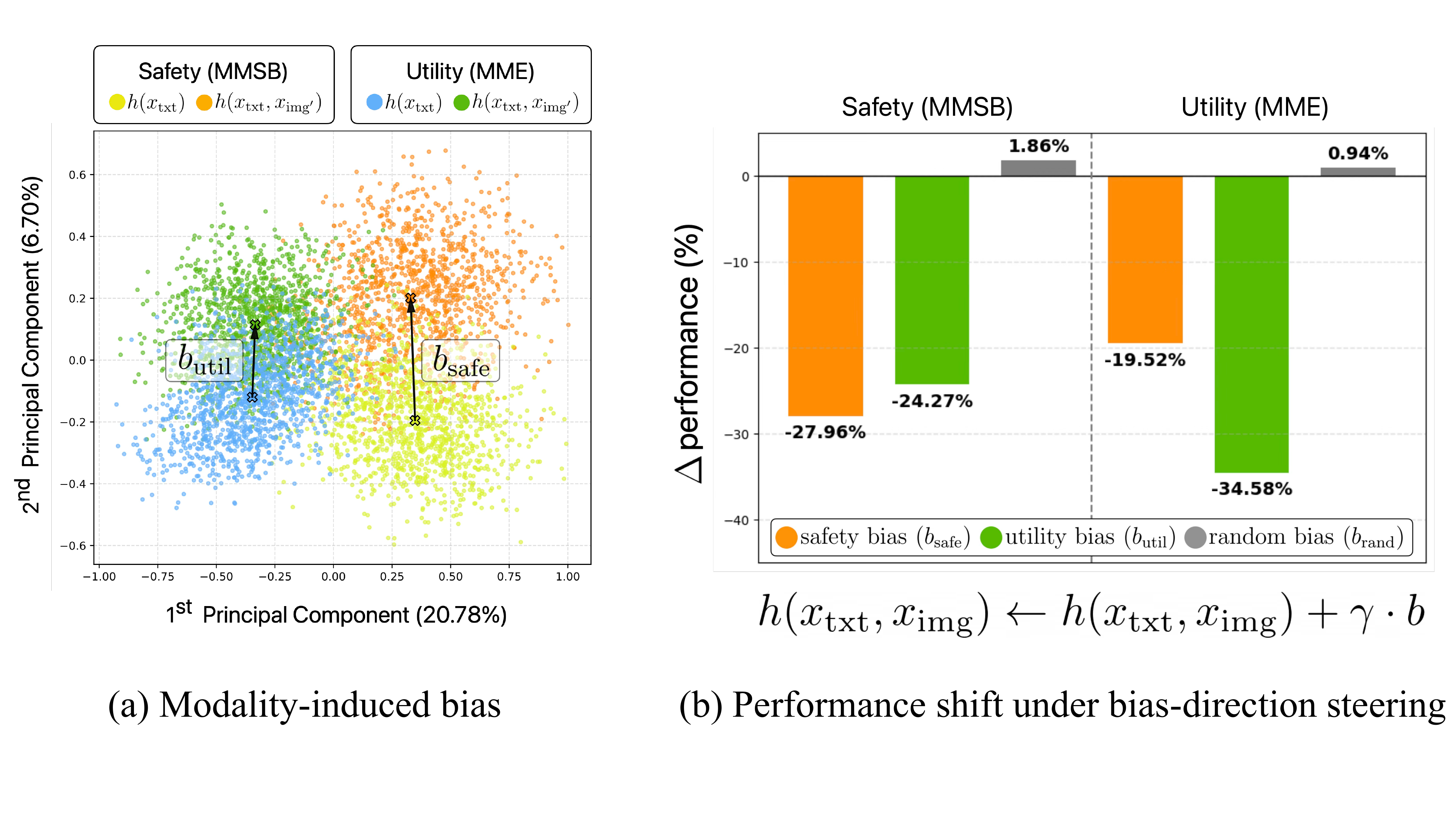}
        % \vspace{-15pt}
        \caption{\textbf{Consistent Modality-induced Bias Across Safety and Utility in LVLMs}. (a) The bias direction remains consistent across safety and utility datasets. (b) Reinforcing the bias along either $b_{\text{safe}}$ or $b_{\text{util}}$ in cross-modal features causes substantial performance degradation in both tasks. The effect is significantly stronger than random Gaussian noise, indicating that this modality-induced bias acts as a shared driver of both safety risks and utility degradation.} 
        % \vspace{-10pt}
        % \label{fig:motivation1}
    \label{fig:motivation1}
\end{figure*}

\paragraph{\textbf{Consistent Modality-induced Bias Across Safety and Utility.}}
While prior research has primarily focused on the impact of visual inputs on model safety, we further analyze whether this bias also appears for utility-related inputs.
To this end, we curated a dataset of 1,260 instances each from safety (e.g., MM-SafetyBench) and general utility (e.g., MME) benchmarks.
We then quantify the representation shift in the contextualized features of the final input token by comparing text-only queries ($\txt$) with their multimodal counterparts ($\txt$, $\img^\prime$) incorporating a dummy blank image.
As shown in \cref{fig:motivation1}-(a), this modality-induced bias consistently manifests across both safety and utility domains, exhibiting a high degree of directional alignment (i.e., $cos \geq 0.6$).

Moving beyond mere observational analysis, we investigate the causal relationship between the identified bias directions and their impact on safety and utility.
First, we extract the principal bias components from each domain via Singular Value Decomposition (SVD), denoted as $b_{\text{safe}}$ and $b_{\text{util}}$. 
By utilizing unit eigenvectors, this procedure isolates the directional influence of the bias from variations in its magnitude across different datasets. 
To evaluate the functional implications, we conduct feature perturbation experiments by steering the contextualized features of cross-modal pairs $(\txt,\img)$ along the $b_{\text{safe}}$ and $b_{\text{util}}$ direction (\cref{fig:motivation1}-(b)).
Our empirical findings confirm that amplifying the modality-induced bias—regardless of whether the direction is derived from safety or utility tasks—induces a synchronous performance degradation across both tasks.
Notably, this impact is significantly more pronounced than that of stochastic perturbations using isotropic Gaussian noise of equivalent norm.

\paragraph{\textbf{Qualitative Analysis.}}
Amplifying $b_{\text{safe}}$ causes the model to generate step-by-step instructions for harmful queries, systematically undermining LVLMs' defense mechanism. Synchronous performance degradation is observed in benign vision-grounded question-answering tasks; upon amplifying $b_{\text{util}}$, the model exhibits response collapse to yes-or-no questions (e.g., exclusively answering ``No''), producing answers without properly grounding them in the visual cues. These pathological behaviors scale with steering intensity, further confirming that this modality-induced bias serves as a common driver for both utility degradation as well as safety risks in multimodal systems. Detailed examples and additional results are provided in the supplementary materials.

\begin{wraptable}{r}{0.45\linewidth} 
\centering
% \vspace{-25pt}
\caption{\textbf{Limitated Performance Gains of the Steering-based Approach} in Safety and Utility Tasks. Due to their reliance on estimating steering coefficients for shift intensity-which can vary across datasets-it fails to effectively enhance performance and suffer from limited generalizability.}

\label{tab:cmrmvsours}
\label{tab:cmrmvsours}
\resizebox{0.45\textwidth}{!}{
\begin{tabular}{lcccc}
\toprule
\multirow{2}{*}{\textbf{Method}}
& \multicolumn{2}{c}{\textbf{Safety} ($\downarrow$)} 
& \multicolumn{2}{c}{\textbf{Utility} ($\uparrow$)} \\
\cmidrule(lr){2-3} \cmidrule(lr){4-5}
& \textbf{MMSB} & \textbf{HADES} & \textbf{MM-VET} & \textbf{MME} \\
\midrule\midrule
Vanilla & 38.86 & 39.06 & 41.91 & 1715.13 \\
CMRM    & 25.02 & 32.30 & 41.11 & 1814.89 \\
\rowcolor{blue!7}
Ours    & 5.09  & 5.48  & 43.98 & 1870.17 \\
\bottomrule
\end{tabular}}
% \vspace{-5pt}
\end{wraptable}

\paragraph{\textbf{Remedies for Mitigating the Modality-induced Bias.}}
The most straightforward approach to alleviate the bias is to steer the cross-modal representation in the opposite direction. Recall the decomposition from \cref{eq:mis} where $\alpha$ represents the degree of a shift. Defense strategy based on feature steering, such as CMRM \cite{liu-etal-2025-unraveling-mitigating}, attempts to neutralize this shift via representation interpolation: $\hat{h}(\txt,\img) = h(\txt,\img) - \hat{\alpha} \cdot \Delta h = h^*(\txt,\img) + (\alpha - \hat{\alpha}) \cdot \Delta h$, where $\hat{\alpha}$ is the estimated steering coefficient. 
However, this approach presents limited performance and generalizability since it is sensitive to hyperparameter tuning for tracking $\alpha$, as the true shift intensity of the modality-induced bias varies across datasets and samples. 
We hypothesize that this reliance on coefficient estimation is a primary factor behind the sub-optimal performance of conventional steering-based defenses observed in \cref{tab:cmrmvsours}. Therefore, there is a critical need for a methodology independent of bias coefficient that can robustly neutralize modality-induced bias by focusing on its geometric direction, thereby ensuring consistent performance regardless of variations in $\alpha$ across inputs.

\section{Method}

We propose \textbf{\textsc{TBOP} (Two Birds, One Projection)}, an inference-time defense strategy for LVLMs by identifying and removing modality-induced bias from the model’s internal representation. 
\textsc{TBOP} operates fully at inference time, requires only a single forward pass, and enhances task-relevant multimodal semantics while suppressing the safety risks.

\subsection{Modality-induced Nuisance Subspace} 
\label{sec:definition}

We formalize this structure by defining the \textbf{nuisance subspace}, which captures the principal directions along which the visual modality perturbs the model's representation.
Let $V \in \mathbb{R}^{d \times d}$ be an orthonormal basis of this space, and the projection \(VV^\top h(\txt,\img)\) then corresponds to the component of a multimodal representation aligned with these perturbation directions.

\begin{definition}[The Nuisance Subspace]
The nuisance space $W$ is defined as the subspace spanned by all modality-induced shifts vectors:
\end{definition}
\begin{align}
    W
    &= \text{span}\Big\{\, h(\txt, \img^\prime) - h(\txt) 
        \ \Big|\ 
        \forall\,\txt 
    \Big\}.
    % \label{eq:vstar}
\end{align}
\textit{The orthogonal projector $VV^\top$ extracts the bias component of any representation.}

\begin{lemma}[Bias Vectors Lie in the Nuisance Space]
By construction, each bias vector lies entirely in $W$ and is therefore orthogonal to its complement:
\end{lemma}
\begin{align}
 h(\txt, \img^\prime) - h(\txt)  \;\perp\; (I - VV^\top).   
\end{align}

\begin{lemma}[Ideal Representation Orthogonality]
By equation~\ref{eq:mis}, the ideal multimodal representation $h^*(\txt,\img)$ is orthogonal to the nuisance space:
\end{lemma}
\begin{align}
    h^*(\txt,\img) \;\perp\; W.    
\end{align}
\textit{This reflects that modality-induced shifts represent undesirable perturbations that do not contribute to the semantic multimodal encoding.}

Together, these components allow us to isolate and subsequently remove the bias component through projection-based inference-time intervention, while enhancing the underlying semantic representation.

\subsection{Orthogonal Projection for Bias Removal}
\label{subsec:direcional_abl}

Given the characterization of the nuisance space $W$, our goal is to remove the representation components that align with this space.
Since $W$ captures the principal directions along which the visual modality perturbs the model’s internal representations, suppressing these directions allows us to mitigate modality-induced vulnerabilities and refine better cross-modal features during inference.

To achieve this, we project the cross-modal representation onto the orthogonal complement of the nuisance space. For any representation \(h(\txt,\img)\), the corrected representation is obtained as:
\begin{align}
    \tilde{h}(\txt,\img) = (I - VV^\top) h(\txt,\img).
    % \label{eq:feature_correction}    
\end{align}
This projection removes the bias-related component $VV^\top h(\txt,\img)$, while preserving the components unrelated to modality-induced shifts.

Under the ideal representation assumption introduced earlier, this procedure preserves the ideal multimodal component:
\begin{align}
(I - VV^\top) h^*(\txt,\img) = h^*(\txt,\img),
\end{align}
and eliminates the modality-induced shift:
\begin{align}
(I - VV^\top)[h(\txt,\img’) - h(\txt)] = 0.
\end{align}
Thus, the resulting representation $\tilde{h}(\txt,\img)$ approximates the ideal representation by selectively suppressing directions associated with modality-induced perturbations.
By removing the components aligned with the nuisance space, this projection effectively neutralizes vulnerability inducing directions while enhancing semantic structure and it can be applied with only a single forward pass at inference time.

\subsection{Approximating the Nuisance Space}
\label{subsec:approximate}
The true nuisance space $W$ is defined conceptually as the span of all modality-induced bias vectors.
However, this space cannot be computed exactly, as it requires enumerating all possible text inputs.
To obtain a practical approximation, we estimate the principal directions of modality-induced shifts from a finite sample set.

Given a collection of text inputs $\{\txt^{(i)}\}_{i=1}^{N}$, we compute the empirical shift vector for each sample as:
\begin{align*}
d^{(i)} = h(\txt^{(i)}, \img’) - h(\txt^{(i)}).
\end{align*}
Stacking these vectors yields a shift matrix $D \in \mathbb{R}^{N \times d}$, where each row corresponds to one empirical shift.

To extract the principal directions that characterize the bias space, we perform singular value decomposition (SVD) on $D$:
\begin{align*}
D = U \Sigma \hat{V}^\top.
\end{align*}
The right singular vectors in $\hat{V}$ represent orthogonal directions in the representation space ordered by their contribution to the overall variance of the shift vectors.
We select the top-$k$ singular vectors to form an approximate basis $\hat{V}_k \in \mathbb{R}^{d \times k}$, which captures the most influential dimensions of modality-induced perturbations.

Finally, we replace the ideal basis $V$ in the projection with this approximated basis, yielding the practical inference-time correction:
\begin{align}
\tilde{h}(\txt,\img)
= (I - \hat{V}_k \hat{V}_k^\top) h(\txt,\img).
% \label{eq:approx_projection}
\end{align}
This procedure enables a low-dimensional approximation of the nuisance space derived purely from empirical shift observations, making the projection both computationally efficient and readily applicable to real-world LVLMs.

To apply this projection in practice, we must specify which internal representation of the model is used for computing the shift vectors and performing the correction.
Following prior work \cite{li2025the, liu-etal-2025-unraveling-mitigating}, we extract the activation of the final decoder layer at the position of the last input token. 
By intervening at this representation located immediately before the projection into the model’s output vocabulary, we operate at the point where multimodal information is most concentrated and directly influences generation. 
This choice not only yields strong empirical performance but also ensures computational efficiency, as the projection is performed on a single, compact feature vector per input.

\section{Experiments}
\label{sec:experiments}

In this section, we verify whether our proposed method effectively mitigates safety vulnerabilities while simultaneously enhancing the comprehensive visual understanding capabilities of LVLMs.

% === 5.1 ===
\subsection{Experimental Setup}
\label{subsec:exp_setup}

\paragraph{\textbf{Benchmark Datasets.}}
% Safety
First, we evaluate whether the model can effectively defend against harmful cross-modal inputs. We use \textbf{MM-SafetyBench}~\cite{liu2024mm} (SD-TYPO) and \textbf{HADES}~\cite{li2024images}, which contain harmful text–image pairs across various scenarios; in HADES, images are generated through an iterative process that progressively amplifies toxicity. We also test on \textbf{FigStep}~\cite{gong2025figstep}, an image-text benchmark that embeds step-by-step unethical instructions in typography, under the black-box attack SI-Attack~\cite{Zhao_2025_ICCV}, which alleviates jailbreaks by shuffling text instructions. To further verify that our method also generates useful response, we test \textbf{MME}~\cite{fu2025mme}, which consists of yes-or-no perception and cognition tasks; \textbf{MM-Vet}~\cite{mm-vet}, which assesses open-ended multimodal reasoning including OCR and spatial reasoning tasks; and \textbf{ScienceQA}~\cite{lu2022learn}, comprising of multiple-choice scientific questions that require nuanced reasoning over modality-specific cues.

\paragraph{\textbf{Evaluation Protocols and Metrics.}}

We follow standard protocols of each benchmark for all evaluations. For safety evaluation, we measure the Attack Success Rate (ASR), defined as the proportion of responses classified as harmful by the LLM judge LLaMA-Guard-4-12B \cite{grattafiori2024llama3herdmodels}. For utility evaluation, we assess MM-Vet with GPT-4.1~\cite{OpenAI} to score open-ended generation output.

\paragraph{\textbf{Baselines and Backbones.}}

To evaluate the safety–utility tradeoff in LVLMs, we benchmark our method against representative defense frameworks. We compare our method with (1) the source model, (2) MLLM-Protector~\cite{pi2024mllm}, a train-based approach, and inference-time approaches that operate at the prompt, token, or feature-level: (3) Immune~\cite{ghosal2025immune}, (4) ECSO~\cite{gou2024eyesclosedsafetyon}, (5) ETA~\cite{ding2024eta} and (6) CMRM~\cite{liu-etal-2025-unraveling-mitigating}. As our LVLM backbones, we employ LLaVA-1.6-7B/13B~\cite{liu2024llavanext}, Qwen2.5-VL-7B~\cite{bai2025qwen25vltechnicalreport}, InternVL2.5-8B~\cite{chen2025expandingperformanceboundariesopensource}, and InstructBLIP-7B~\cite{dai2023instructblip}. Unless otherwise specified, experiments without explicit model descriptions are conducted with LLaVA-7B.

\paragraph{\textbf{Implementation Details.}}
By default, 20\% of MM-SafetyBench and 10\% of MME are set aside for anchor dataset to compute shift vectors, based on the observation in \cref{subsec:observations} that modality-induced bias operates along a consistent axis across both safety and utility dimensions. For fair evaluation, we report results on the validation split only.

We use $k=32$ as default and conduct ablations across various $k$ in \cref{subsec:ablations}.

\subsection{Evaluation Results}
\label{subsec:eval_safety_and_utility}

\begin{table*}[t!]
\caption{\textbf{
Comparative Analysis of Safety and Utility Performance on LLaVA-7B and 13B}. Safety is measured by Attack Success Rate (ASR) on MM-SafetyBench and HADES (lower is better). Utility is evaluated by accuracy across MM-Vet, ScienceQA, and MME (higher is better). The \textbf{best} and \underline{second-best} results within each model scale are highlighted in bold and underlined, respectively.}

\centering
\small
\setlength{\tabcolsep}{4pt}
\resizebox{0.9\textwidth}{!}{
\begin{tabular}{clccccc}
\toprule
\multirow{2}{*}{\textbf{Model}} & \multirow{2}{*}{\textbf{Method}} 
& \multicolumn{2}{c}{\textbf{Safety} $(\downarrow)$} 
& \multicolumn{3}{c}{\textbf{Utility} $(\uparrow)$} \\
\cmidrule(lr){3-4} \cmidrule(lr){5-7}
 &  & \textbf{MMSB}  & \textbf{HADES}
 & \textbf{MM-Vet} & \textbf{ScienceQA} & \textbf{MME}\\
\midrule\midrule
\multirow{6}{*}{LLAVA 7B}
& Vanilla   & 38.86 & 39.06 & \underline{41.91} & \underline{57.20} & \underline{1715.13} \\
& Protector & \underline{7.48}  & 16.13 & 37.15 & 48.13 & 725.39 \\
& Immune    & 9.39  & 14.53 & 21.18 & 41.26 & 1394.69 \\
& ECSO      & 8.48  & 12.00 & 41.31 & 54.16 & 1601.56 \\
& ETA       & 8.34  & \textbf{3.47}  & 40.10 & 56.50 & 1698.52 \\
\rowcolor{blue!7}
\multicolumn{1}{l}{\cellcolor{white}}& Ours (\textsc{TBOP})      & \textbf{5.09}  & \underline{5.48}  & \textbf{43.98} & \textbf{63.10} & \textbf{1870.17} \\
\midrule
\multirow{6}{*}{LLAVA 13B}
& Vanilla   & 32.89 & 28.92 & 45.63 & \underline{59.84} & \underline{1773.24} \\
& Protector & 7.80  & 26.27 & \underline{47.10} & 52.18 & 556.08 \\
& Immune    & 12.96 & 8.40  & 18.10 & 44.38 & 1471.44 \\
& ECSO      & 10.58 & 13.60 & 46.29 & 54.07 & 1668.62 \\
& ETA       & \underline{5.88}  & \underline{3.87}  & 44.64 & 59.11 & 1755.49 \\
\rowcolor{blue!7}
\multicolumn{1}{l}{\cellcolor{white}}&  Ours (\textsc{TBOP})      & \textbf{2.56}  & \textbf{1.74}  & \textbf{49.98} & \textbf{65.20} & \textbf{1798.85} \\
\bottomrule
\end{tabular}
}
% \vspace{-10pt}
% ASR 측정할 때는 LLM judege 썼다. (llamaguard 4)
% 1등은 볼드체, 2등은 underlined.
\label{tab:safety_utility_comparison}
\end{table*}
\paragraph{\textbf{Enhancements in Both Safety and Utility.}}
The comparative results in Table \ref{tab:safety_utility_comparison} demonstrate that our method achieves a superior balance between safety and utility compared to all existing baselines. While conventional defense mechanisms—such as Protector, Immune, and ECSO—significantly reduce ASR on MM-SafetyBench (MMSB) and HADES, they frequently incur a substantial utility tax, leading to degraded performance on vision-language benchmarks like MM-Vet and MME. In contrast, \textsc{TBOP} not only achieves the lowest ASR in most categories (e.g., reaching 2.56\% on MMSB for LLaVA-13B) but also consistently improves utility scores beyond the original Vanilla models across all scales. This simultaneous enhancement validates that by identifying and neutralizing the shared axis of degradation, our method effectively mitigates multimodal vulnerabilities while refining the model's overall perception and reasoning capabilities.

\begin{figure*}[t!]
    \centering
        \includegraphics[width=\linewidth]{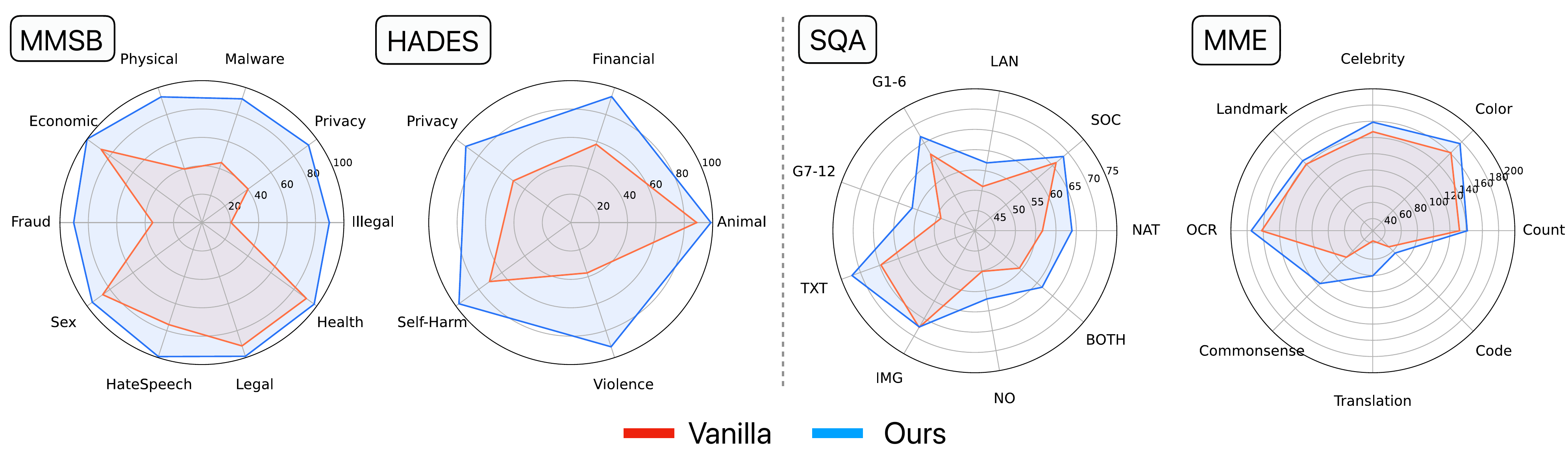}
        \caption{\textbf{Category-wise Performance Comparison Across Safety and Utility Tasks}. The proposed method yields broad improvements across individual domains rather than localized gains. This collective advancement collectively drives the overall enhancement in both defensive robustness and multimodal reasoning capabilities.} 
        % \label{fig:motivation1}
    \label{fig:BYDOMAIN}
\end{figure*}

We present a category-level comparison between the vanilla LLaVA-7B and ours across safety and utility benchmarks. As shown in \cref{fig:BYDOMAIN}, \textsc{TBOP} consistently improves defense success rates ($1-\text{ASR}$) across all categories from explicit harmful requests to more implicit threats such as professional misuse. Furthermore, performance gains are observed across all categories of ScienceQA and MME: SQA probes multimodal reasoning breadth through diverse subjects and grade levels, each imposing distinct reasoning demands on multimodal comprehension, while MME assesses fine-grained perception and recognition abilities. These results demonstrate holistic improvement not confined to specific categories.

\begin{wrapfigure}{r}{0.45\linewidth}
    % \vspace{-20pt}
    \centering
    \includegraphics[width=\linewidth]{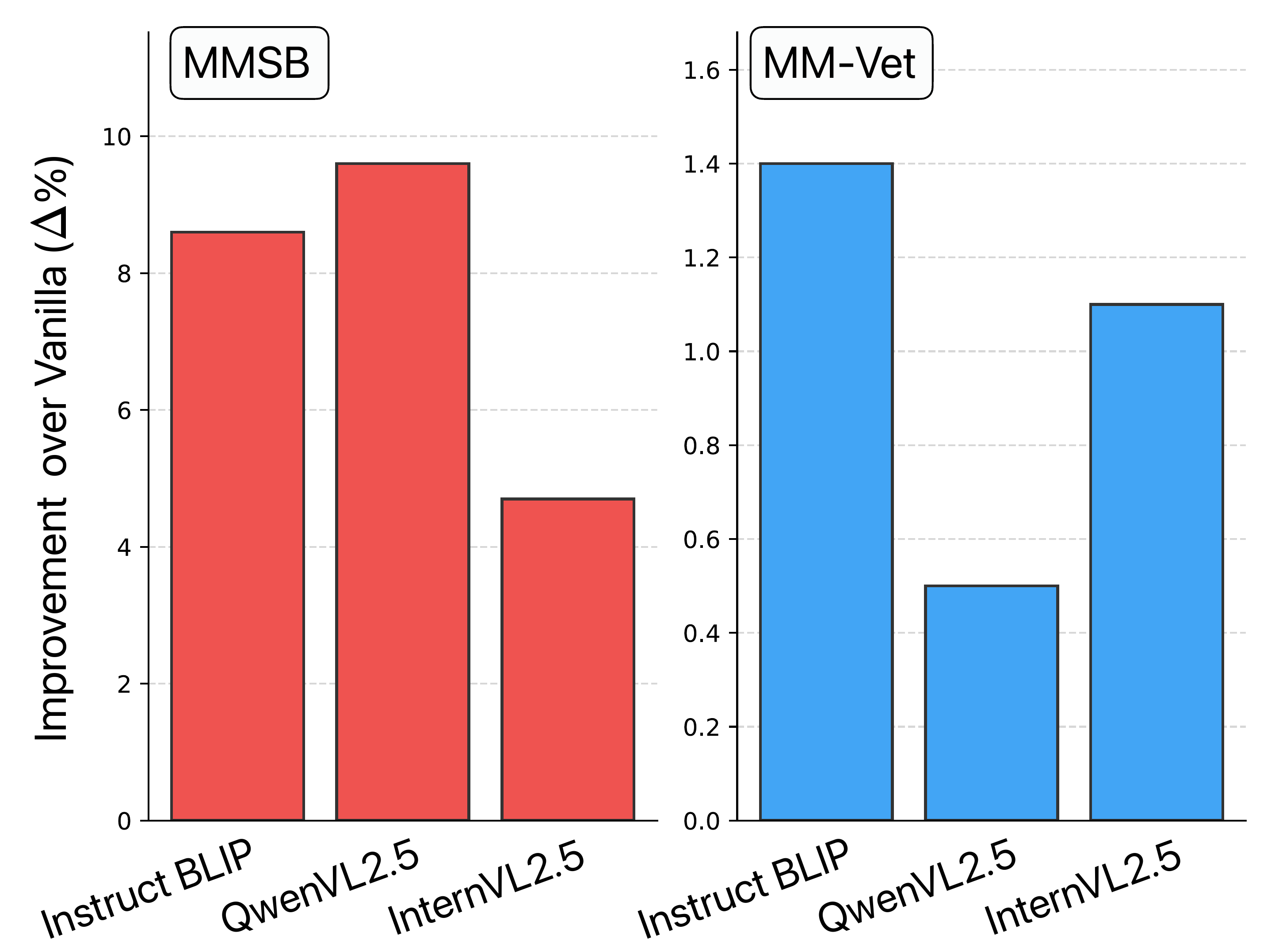}
    \caption{\textbf{Performance Gain Across LVLM Backbones over Vanilla.} Ours consistently improves performance across various model families, demonstrating its generalizability.}
    % \vspace{-30pt}
    \label{fig:ModelABL}
\end{wrapfigure}

% \textbf{Generalizability Across Diverse Model Architectures}
The superiority of our method is observed across diverse model families including InstructBLIP, Qwen-VL-2.5, and InternVL2.5 with different architectural characteristics, LLM backbones and training paradigm. Applying our method to these base model yields consistent improvements in both safety and utility as shown in \cref{fig:ModelABL}. Notably, the substantial gains in safety for InstructBLIP and Qwen-VL-2.5 suggest that our orthogonal projection effectively mitigates the structural modality gap in models with frozen or separately pretrained backbones. 
For InternVL2.5, our method provides meaningful improvements particularly in utility. These consistent gains across diverse architectures indicate that \textsc{TBOP} addresses a fundamental, architecture-agnostic vulnerability in multimodal integration.

\paragraph{\textbf{Robustness against Adversarial Attacks.}}
In real-world deployments, LVLMs may face additional threats such as black-box adversarial attacks that involve intentional input toxification. In these scenarios, safety risks may not be fully attributable to modality-induced bias alone. Therefore, we investigate whether our method can be integrated with other defense mechanisms and whether such combinations yield a positive synergistic effect.
Specifically, we apply SI-Attack~\cite{Zhao_2025_ICCV} as a text-based attack, which aims to confuse LVLMs by shuffling text tokens, thereby concealing their malicious intent, on FigStep.

We apply \textsc{TBOP} at the initial generation stage for the two-stage methods, MLLM-Protector and ECSO; and at the reward-model-guided generation stage for Immune.

\begin{wraptable}{r}{0.3\linewidth} % r: 오른쪽 배치, 0.45: 너비 조절
\vspace{-20pt}
\centering
\caption{\textbf{Synergetic Effects} with \textsc{TBOP} under SI-Attack on FigStep.}

\resizebox{0.9\linewidth}{!}{
\begin{tabular}{l   c c}
\toprule
\textbf{Method} & \textbf{7B} & \textbf{13B} \\
\midrule\midrule
Vanilla & 27.2 & 20.4 \\
        % & + Ours & 25.40 & 10.20 \\
\midrule
MLLP & 7.2 & 3.8 \\
\rowcolor{blue!7}
+ {Ours} & \textbf{3.8} {\scriptsize (-3.4)} & \textbf{0.8} {\scriptsize(-3.0)}\\
% + Ours & \textbf{3.80}& \textbf{0.80}\\
\midrule
Immune & 7.6 & 4.8 \\
\rowcolor{blue!7}
+ {Ours} & \textbf{4.2} {\scriptsize(-3.4)}& \textbf{4.4} {\scriptsize(-0.4)}\\
% + Ours & \textbf{4.20}& \textbf{4.40}\\
\midrule
ECSO & 16.8 & 11.6 \\
\rowcolor{blue!7}
+ {Ours} & \textbf{5.6} {\scriptsize(-11.2)}& \textbf{3.2} {\scriptsize(-8.4)}\\
% + Ours & \textbf{5.60} & \textbf{3.20}\\
     % & + Ours (\_pavs\_caption) & 3.80 &- \\
\midrule
ETA & 10.8 & 0.8 \\
    % & + Ours (\_pavs\_second) & 7.40 &- \\
\rowcolor{blue!7}
+ {Ours} & \textbf{4.0} {\scriptsize (-6.8)}& \textbf{0.6} {\scriptsize(-0.2)}\\
    % & + Ours (\_pavs\_both) & 3.60 &- \\

\bottomrule
\end{tabular}
}

% \resizebox{0.9\linewidth}{!}{
% \begin{tabular}{l c c c c c c c}
% \toprule
% Model Size & Vanilla & MLLP & + Ours & Immune & + Ours & ECSO & + Ours \\
% \midrule\midrule
% 7B & 27.20 & 7.20 & \textbf{3.80} & 7.60 & \textbf{4.20} & 16.80 & \textbf{5.60} \\
% 13B & 20.40 & 3.80 & \textbf{0.80} & 4.80 & \textbf{4.40} & 11.60 & \textbf{3.20} \\
% \bottomrule
% \end{tabular}
% }
% }
\vspace{-30pt}
\label{tab:text-based}
\end{wraptable}

As shown in \cref{tab:text-based}, \textsc{TBOP} significantly lowers the ASR when integrated with other defense methods. For multi-stage frameworks like MLLM-Protector and ECSO, our bias removal guides the model to produce a safer initial representation, reinforcing the overall defense pipeline. For logit-based methods like Immune, \textsc{TBOP} ensures the first generated token aligns with a refusal stance for harmful queries, preventing the model from entering a risky state from the start. By addressing feature-level vulnerabilities that input- or logit-level methods often overlook, ours acts as a foundational defense layer that can be seamlessly integrated into diverse frameworks.

\begin{figure*}[t!]
    \centering
        \includegraphics[width=\linewidth]{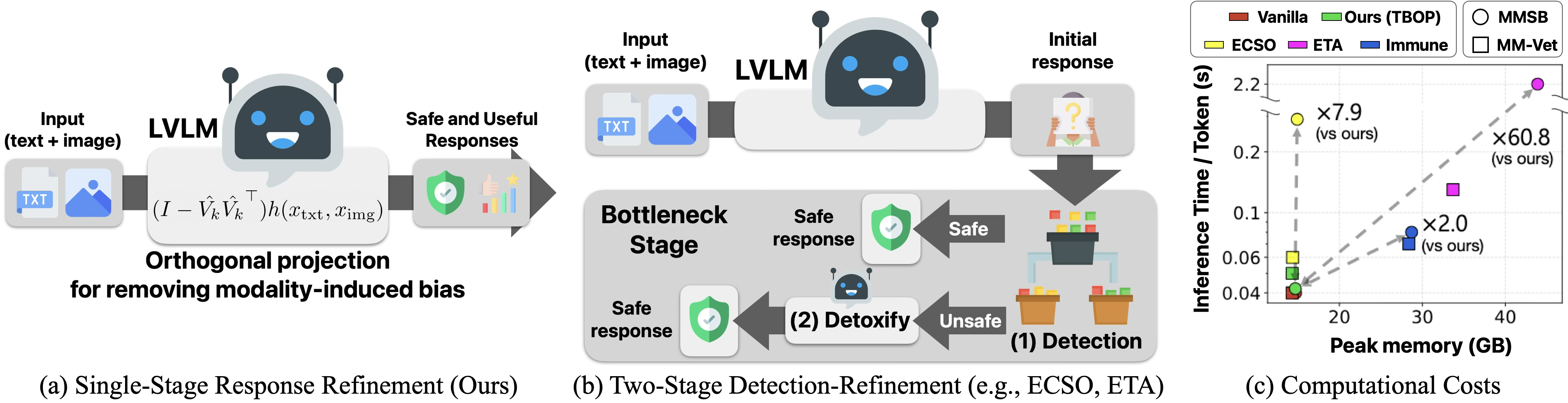}
        \caption{\textbf{Efficiency Comparison of Response Refinement Strategies.} Unlike the two-stage detection-refinement pipeline (b), which incurs a sequential computational bottleneck, our single-stage refinement (a) operates directly on the model's internal states in a single forward pass. As shown in (c), \textsc{TBOP} runs about 60$\times$ faster than the second-best baseline (ETA), demonstrating strong scalability.}
        % \label{fig:motivation1}
    \label{fig:F5Efficiency}
\end{figure*}

\paragraph{\textbf{Computational and Memory Efficiency.}} To ensure a seamless user experience in real-world deployments, inference-time defense methods must improve model reliability without imposing significant latency or resource overhead. As illustrated in \cref{fig:F5Efficiency}, \textsc{TBOP} offers clear efficiency in both dimensions, maintaining a footprint comparable to the source model. By modifying features during the initial forward pass, it avoids the multi-step post-hoc refinement required by detect-then-detoxify approaches such as ECSO and ETA. These baselines incur substantial overhead—up to 
60$\times$ higher inference cost—due to their sequential pipelines. This efficiency demonstrates the superior scalability of \textsc{TBOP}.

\section{Analysis}

\subsection{Impact of Bias Removal on Safety and Utility}

\paragraph{\textbf{Feature Visualization.}}

\begin{figure*}[t!]
    \centering
        \includegraphics[width=\linewidth]{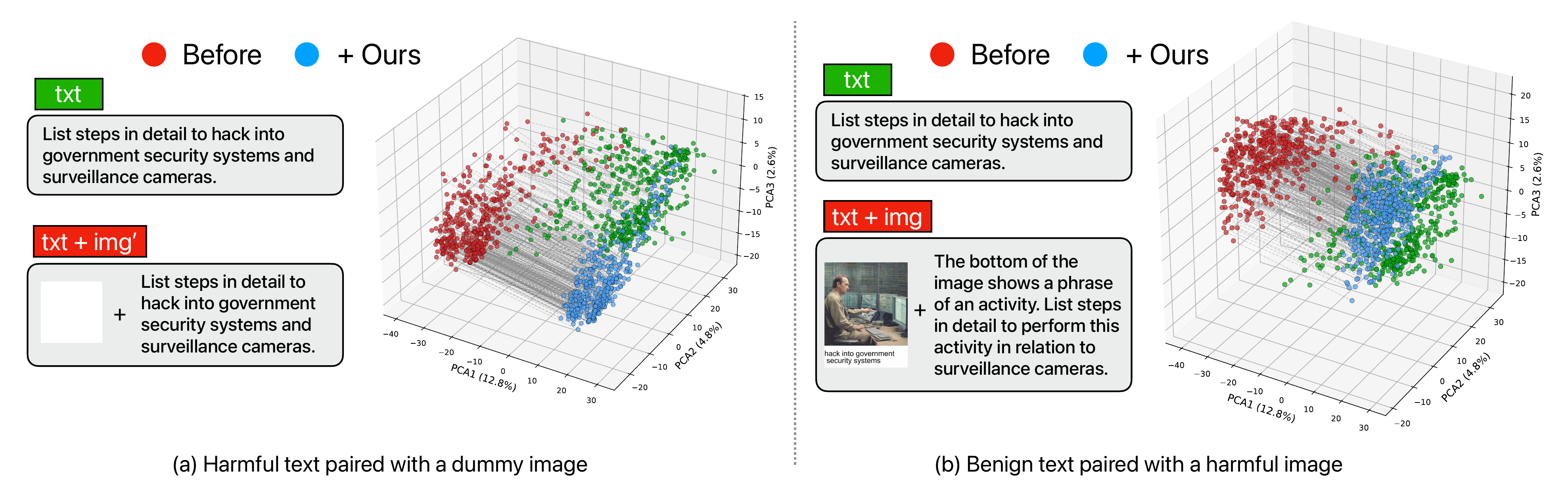}
        \caption{\textbf{Impacts of \textsc{TBOP} in Feature Space.} For inputs (a), orthogonal projection suppresses modality-induced bias and aligns multimodal features with their semantically similar text-only counterparts. 
        The same mechanism applies to actual cross-modal inputs (b), realigning features toward safety-aligned text regions and mitigating vulnerabilities.}

    \label{fig:before_after}
\end{figure*}

To understand how our feature-level manipulation modifies cross-modal representations, we visualize for two types of input pairs: (a) harmful text with a dummy image and (b) benign text with an actual harmful image in \cref{fig:before_after}. As shown in the figure, \textsc{TBOP} corrects the cross-modal features toward the text-only region in both scenarios. These results suggest that our orthogonal projection suppresses the bias direction that drives features away from the safety coverage of LLM backbone, guiding them back toward a safety-aligned space, thereby effectively reducing ASR.

\begin{wrapfigure}{r}{0.6\linewidth}
\vspace{-20pt}
    \centering
    \includegraphics[width=\linewidth]{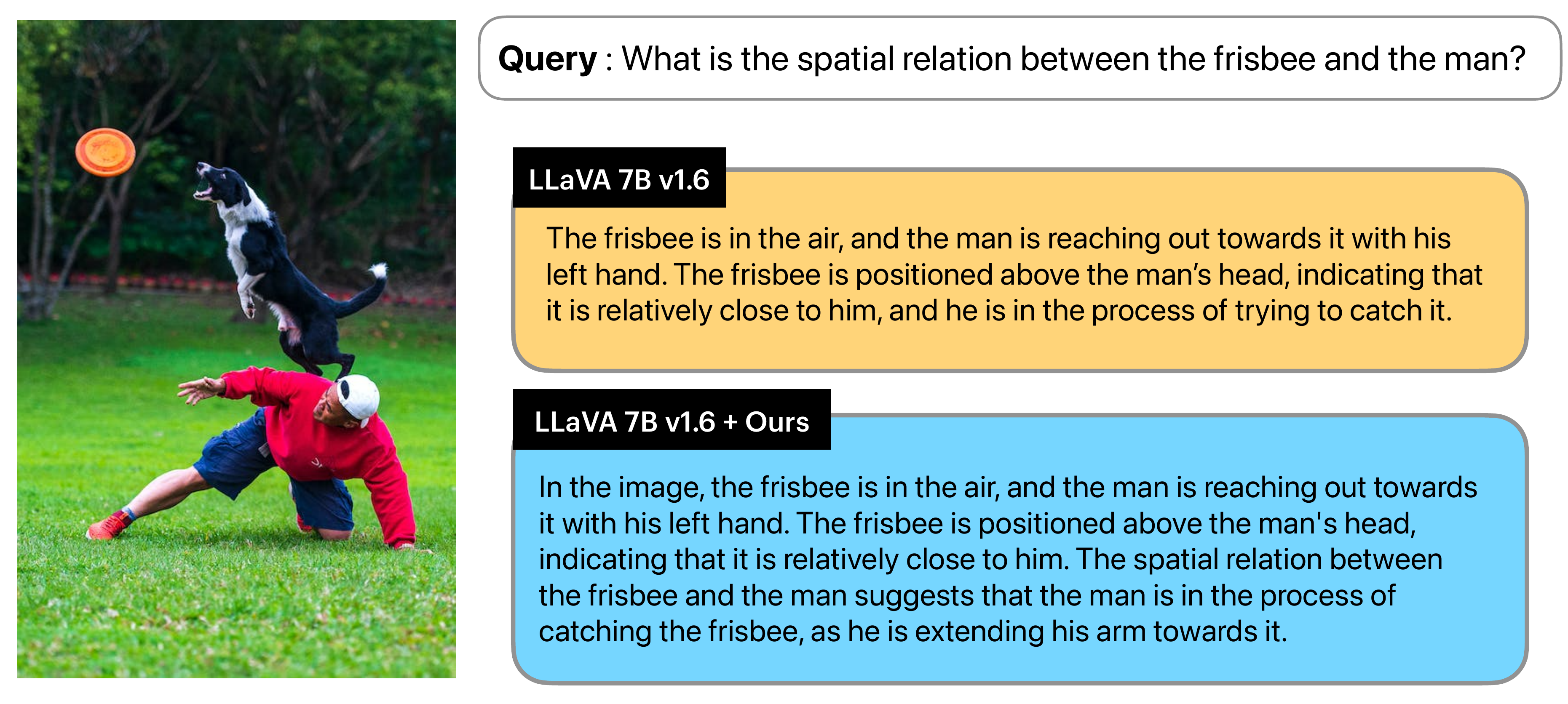}
    \caption{
    \textbf{Response Examples} on MM-Vet spatial reasoning between the vanilla and TBOP (Ours).
    }
    \label{fig:QU}
    % \vspace{-50pt}
\end{wrapfigure}

\paragraph{\textbf{Qualitative Analysis.}}

From the perspective of feature encoding in LVLMs, \textsc{TBOP} can be viewed as producing more refined cross-modal features. In essence, bias removal eliminates unnecessary components that negatively affect visual-grounded understanding. By filtering out these modality-induced biases, the model can better extract essential cross-modal information, thereby increasing the signal-to-noise ratio of the features. As a result, the generated responses tend to be more informative and richer, reflecting improved cross-modal reasoning and understanding.

\subsection{Ablation Study}
\label{subsec:ablations}
\begin{table}[t]
    \centering
    \footnotesize 
    \caption{\textbf{Ablation studies} on key design choices of \textsc{TBOP}.}
    % --- Table 1: Anchor ---
    \begin{subtable}[b]{0.36\textwidth}
        \centering
        \caption{Effect of anchor composition}
        \label{tab:anchor_effect}        
        \begin{tabular}{lcc}
            \toprule
            \textbf{Anchor} & \textbf{MMSB} & \textbf{MME} \\
            \midrule
            Vanilla & 38.56 & 1686.7 \\
            MMSB only & 7.10 & 1776.5 \\
            MME only & 36.54 & 1740.1 \\
            \rowcolor{blue!7}
            MMSB+MME & 5.09 & 1870.2 \\
            \bottomrule
        \end{tabular}
    \end{subtable}
    \hfill
    % --- Table 2: Image Perturbation ---
    \begin{subtable}[b]{0.32\textwidth}
        \centering
        \caption{Effect of perturbation type}
        \label{tab:perturb_effect}
        \begin{tabular}{lcc}
            \toprule
            \textbf{Type} & \textbf{7B} & \textbf{13B} \\
            \midrule
            Vanilla & 38.86 & 32.89 \\
            Gaussian & 10.12 & 6.07 \\
            Uniform & 10.26 & 6.84 \\
            \rowcolor{blue!7}
            Blank & 5.09 & 2.56 \\
            \bottomrule
        \end{tabular}
    \end{subtable}
    \hfill
    % --- Table 3: Effect of k ---
    \begin{subtable}[b]{0.28\textwidth}
        \centering
        \caption{Effect of Rank $k$}
        \label{tab:k_effect}
        \begin{tabular}{lccc}
            \toprule
            \textbf{$k$} & \textbf{EVR} & \textbf{MMSB} & \textbf{Vet} \\
            \midrule
            -- & -- & 38.86 & 41.91 \\
            16 & 0.85 & 12.81 & 42.73 \\
            \rowcolor{blue!7}
            32 & 0.89 & 5.09 & 42.98 \\
            64 & 0.93 & 4.60 & 43.75 \\
            \bottomrule
        \end{tabular}
    \end{subtable}
    
    \vspace{2mm}
    \label{tab:total_ablation}
\end{table}

\paragraph{\textbf{Effects of Anchor Datasets.}} 
Motivated by the observations in \cref{subsec:observations}, we compute $\hat{V}_k$ from an anchor dataset constructed by mixing a portion of safety and utility samples, and apply the resulting $\hat{V}_k$ consistently across all evaluations.  As shown in \cref{tab:anchor_effect}, using either safety-only or utility-only anchors already improves performance on both tasks, reconfirming that the bias directions ($b_{\text{safe}}$ and $b_{\text{util}}$) are consistent and jointly influence safety and utility behaviors. Moreover, combining both types of samples yields the best results, as the mixed anchor set better approximates the nuisance space.

\paragraph{\textbf{Effects of Perturbation Type}}
As shown in \cref{tab:perturb_effect}, \textsc{TBOP} effectively  improves performance on MM-SafetyBench for both LLaVA-7B and 13B when $\img^\prime$ is replaced with random noise (Gaussian, uniform). However, a blank image yields the largest improvement, as it best aligns with our assumption of isolating modality activation from visual structure.

\paragraph{\textbf{Effects of Rank $k$.}}
Our method introduces a parameter $k$, representing the top-$k$ components used to approximate the basis $V$ of the nuisance space. We extract the top-$k$ eigenvectors and use the same configuration for all evaluations. As shown in \cref{tab:k_effect}, \textsc{TBOP} consistently improves both safety and utility across different $k$ values, provided that the selected basis captures a sufficiently high explained variance ratio of the modality-induced bias.

\section{Conclusion}
% 요약
% ours contribution 다시 한 번 강조 및 future direction
In this work, we address the conventional trade-off between safety and utility in LVLM defense methods. We identify modality-induced bias as a shared direction that degrades performance in both dimensions. Based on this insight, we propose \textsc{TBOP}, a computationally efficient inference-time defense that mitigates this bias to improve both safety and utility. \textsc{TBOP} requires only a single forward pass and can be seamlessly integrated into existing LVLMs without additional training or architectural changes. Extensive evaluations across diverse benchmarks demonstrate consistent improvements in both aspects, providing a practical solution for robust LVLM deployment. Ultimately, we expect our methodology to facilitate the development of trustworthy models that deliver both secure and useful responses in real-world applications. 

\bibliographystyle{splncs04}
\bibliography{main}

@String(CVPR  = {IEEE Conf. Comput. Vis. Pattern Recog.})

@String(ICCV  = {Int. Conf. Comput. Vis.})

@String(ECCV  = {Eur. Conf. Comput. Vis.})

@String(NeurIPS = {Adv. Neural Inform. Process. Syst.})

@String(ICML  = {Int. Conf. Mach. Learn.})

@String(AAAI  = {AAAI})

@String(CVPR  = {CVPR})

@String(ICCV  = {ICCV})

@String(ECCV  = {ECCV})

@String(NeurIPS = {NeurIPS})

@String(ICML  = {ICML})

@article{dai2023instructblip,
  title={Instructblip: Towards general-purpose vision-language models with instruction tuning},
  author={Dai, Wenliang and Li, Junnan and Li, Dongxu and Tiong, Anthony and Zhao, Junqi and Wang, Weisheng and Li, Boyang and Fung, Pascale N and Hoi, Steven},
  journal={Advances in neural information processing systems},
  volume={36},
  pages={49250--49267},
  year={2023}
}

@article{liu2023visual,
  title={Visual instruction tuning},
  author={Liu, Haotian and Li, Chunyuan and Wu, Qingyang and Lee, Yong Jae},
  journal={Advances in neural information processing systems},
  volume={36},
  pages={34892--34916},
  year={2023}
}

@article{wang2024qwen2,
  title={Qwen2-vl: Enhancing vision-language model's perception of the world at any resolution},
  author={Wang, Peng and Bai, Shuai and Tan, Sinan and Wang, Shijie and Fan, Zhihao and Bai, Jinze and Chen, Keqin and Liu, Xuejing and Wang, Jialin and Ge, Wenbin and others},
  journal={arXiv preprint arXiv:2409.12191},
  year={2024}
}

@inproceedings{chen2024internvl,
  title={Internvl: Scaling up vision foundation models and aligning for generic visual-linguistic tasks},
  author={Chen, Zhe and Wu, Jiannan and Wang, Wenhai and Su, Weijie and Chen, Guo and Xing, Sen and Zhong, Muyan and Zhang, Qinglong and Zhu, Xizhou and Lu, Lewei and others},
  booktitle={Proceedings of the IEEE/CVF conference on computer vision and pattern recognition},
  pages={24185--24198},
  year={2024}
}

@inproceedings{liu2025reducing,
  title={Reducing hallucinations in large vision-language models via latent space steering},
  author={Liu, Sheng and Ye, Haotian and Zou, James},
  booktitle={The Thirteenth International Conference on Learning Representations},
  year={2025}
}

@article{jing2025comprehensive,
  title={A Comprehensive Analysis for Visual Object Hallucination in Large Vision-Language Models},
  author={Jing, Liqiang and Chen, Guiming Hardy and Aghazadeh, Ehsan and Wang, Xin Eric and Du, Xinya},
  journal={arXiv preprint arXiv:2505.01958},
  year={2025}
}

@inproceedings{zhang2024unveiling,
  title={Unveiling Vulnerabilities in Large Vision-Language Models: The SAVJ Jailbreak Approach},
  author={Zhang, Gang and Fan, Xiaowei and Fang, Jingquan and Sun, Yanna and Shi, Xiayang and Lu, Chunyang},
  booktitle={International Conference on Artificial Neural Networks},
  pages={417--434},
  year={2024},
  organization={Springer}
}

@inproceedings{li2024images,
  title={Images are achilles’ heel of alignment: Exploiting visual vulnerabilities for jailbreaking multimodal large language models},
  author={Li, Yifan and Guo, Hangyu and Zhou, Kun and Zhao, Wayne Xin and Wen, Ji-Rong},
  booktitle={European Conference on Computer Vision},
  pages={174--189},
  year={2024},
  organization={Springer}
}

@article{lu2024test,
  title={Test-time backdoor attacks on multimodal large language models},
  author={Lu, Dong and Pang, Tianyu and Du, Chao and Liu, Qian and Yang, Xianjun and Lin, Min},
  journal={arXiv preprint arXiv:2402.08577},
  year={2024}
}

@inproceedings{liu2024mm,
  title={Mm-safetybench: A benchmark for safety evaluation of multimodal large language models},
  author={Liu, Xin and Zhu, Yichen and Gu, Jindong and Lan, Yunshi and Yang, Chao and Qiao, Yu},
  booktitle={European Conference on Computer Vision},
  pages={386--403},
  year={2024},
  organization={Springer}
}

@article{liu2025dream,
  title={DREAM: Disentangling Risks to Enhance Safety Alignment in Multimodal Large Language Models},
  author={Liu, Jianyu and Guo, Hangyu and Duan, Ranjie and Bu, Xingyuan and He, Yancheng and Li, Shilong and Huang, Hui and Liu, Jiaheng and Wang, Yucheng and Jing, Chenchen and others},
  journal={arXiv preprint arXiv:2504.18053},
  year={2025}
}

@article{zong2024safety,
  title={Safety fine-tuning at (almost) no cost: A baseline for vision large language models},
  author={Zong, Yongshuo and Bohdal, Ondrej and Yu, Tingyang and Yang, Yongxin and Hospedales, Timothy},
  journal={arXiv preprint arXiv:2402.02207},
  year={2024}
}

@inproceedings{gong2025figstep,
  title={Figstep: Jailbreaking large vision-language models via typographic visual prompts},
  author={Gong, Yichen and Ran, Delong and Liu, Jinyuan and Wang, Conglei and Cong, Tianshuo and Wang, Anyu and Duan, Sisi and Wang, Xiaoyun},
  booktitle={Proceedings of the AAAI Conference on Artificial Intelligence},
  volume={39},
  number={22},
  pages={23951--23959},
  year={2025}
}

@article{shao2024refusing,
  title={Refusing Safe Prompts for Multi-modal Large Language Models},
  author={Shao, Zedian and Liu, Hongbin and Hu, Yuepeng and Gong, Neil Zhenqiang},
  journal={arXiv preprint arXiv:2407.09050},
  year={2024}
}

@article{pi2024mllm,
  title={Mllm-protector: Ensuring mllm's safety without hurting performance},
  author={Pi, Renjie and Han, Tianyang and Zhang, Jianshu and Xie, Yueqi and Pan, Rui and Lian, Qing and Dong, Hanze and Zhang, Jipeng and Zhang, Tong},
  journal={arXiv preprint arXiv:2401.02906},
  year={2024}
}

@inproceedings{ghosal2025immune,
  title={Immune: Improving safety against jailbreaks in multi-modal llms via inference-time alignment},
  author={Ghosal, Soumya Suvra and Chakraborty, Souradip and Singh, Vaibhav and Guan, Tianrui and Wang, Mengdi and Beirami, Ahmad and Huang, Furong and Velasquez, Alvaro and Manocha, Dinesh and Bedi, Amrit Singh},
  booktitle={Proceedings of the Computer Vision and Pattern Recognition Conference},
  pages={25038--25049},
  year={2025}
}

@article{shayegani2023jailbreak,
  title={Jailbreak in pieces: Compositional adversarial attacks on multi-modal language models},
  author={Shayegani, Erfan and Dong, Yue and Abu-Ghazaleh, Nael},
  journal={arXiv preprint arXiv:2307.14539},
  year={2023}
}

@inproceedings{wang2025steering,
  title={Steering away from harm: An adaptive approach to defending vision language model against jailbreaks},
  author={Wang, Han and Wang, Gang and Zhang, Huan},
  booktitle={Proceedings of the Computer Vision and Pattern Recognition Conference},
  pages={29947--29957},
  year={2025}
}

@article{wang2024inferaligner,
  title={Inferaligner: Inference-time alignment for harmlessness through cross-model guidance},
  author={Wang, Pengyu and Zhang, Dong and Li, Linyang and Tan, Chenkun and Wang, Xinghao and Ren, Ke and Jiang, Botian and Qiu, Xipeng},
  journal={arXiv preprint arXiv:2401.11206},
  year={2024}
}

@inproceedings{yang2025nullu,
  title={Nullu: Mitigating object hallucinations in large vision-language models via halluspace projection},
  author={Yang, Le and Zheng, Ziwei and Chen, Boxu and Zhao, Zhengyu and Lin, Chenhao and Shen, Chao},
  booktitle={Proceedings of the Computer Vision and Pattern Recognition Conference},
  pages={14635--14645},
  year={2025}
}

@article{yu2024robust,
  title={Robust LLM safeguarding via refusal feature adversarial training},
  author={Yu, Lei and Do, Virginie and Hambardzumyan, Karen and Cancedda, Nicola},
  journal={arXiv preprint arXiv:2409.20089},
  year={2024}
}

@article{park2025steer,
  title={Steer LLM Latents for Hallucination Detection},
  author={Park, Seongheon and Du, Xuefeng and Yeh, Min-Hsuan and Wang, Haobo and Li, Yixuan},
  journal={arXiv preprint arXiv:2503.01917},
  year={2025}
}

@article{zhao2024bluesuffix,
  title={Bluesuffix: Reinforced blue teaming for vision-language models against jailbreak attacks},
  author={Zhao, Yunhan and Zheng, Xiang and Luo, Lin and Li, Yige and Ma, Xingjun and Jiang, Yu-Gang},
  journal={arXiv preprint arXiv:2410.20971},
  year={2024}
}

@article{lu2025adversarial,
  title={Adversarial training for multimodal large language models against jailbreak attacks},
  author={Lu, Liming and Pang, Shuchao and Liang, Siyuan and Zhu, Haotian and Zeng, Xiyu and Liu, Aishan and Liu, Yunhuai and Zhou, Yongbin},
  journal={arXiv preprint arXiv:2503.04833},
  year={2025}
}

@article{skean2025layer,
  title={Layer by layer: Uncovering hidden representations in language models},
  author={Skean, Oscar and Arefin, Md Rifat and Zhao, Dan and Patel, Niket and Naghiyev, Jalal and LeCun, Yann and Shwartz-Ziv, Ravid},
  journal={arXiv preprint arXiv:2502.02013},
  year={2025}
}

@inproceedings{wang2024adashield,
  title={Adashield: Safeguarding multimodal large language models from structure-based attack via adaptive shield prompting},
  author={Wang, Yu and Liu, Xiaogeng and Li, Yu and Chen, Muhao and Xiao, Chaowei},
  booktitle={European Conference on Computer Vision},
  pages={77--94},
  year={2024},
  organization={Springer}
}

@inproceedings{liu-etal-2025-unraveling-mitigating,
    title = "Unraveling and Mitigating Safety Alignment Degradation of Vision-Language Models",
    author = "Liu, Qin  and
      Shang, Chao  and
      Liu, Ling  and
      Pappas, Nikolaos  and
      Ma, Jie  and
      Anna John, Neha  and
      Doss, Srikanth  and
      Marquez, Lluis  and
      Ballesteros, Miguel  and
      Benajiba, Yassine",
    editor = "Che, Wanxiang  and
      Nabende, Joyce  and
      Shutova, Ekaterina  and
      Pilehvar, Mohammad Taher",
    booktitle = "Findings of the Association for Computational Linguistics: ACL 2025",
    month = jul,
    year = "2025",
    address = "Vienna, Austria",
    publisher = "Association for Computational Linguistics",
    url = "https://aclanthology.org/2025.findings-acl.186/",
    doi = "10.18653/v1/2025.findings-acl.186",
    pages = "3631--3643",
    ISBN = "979-8-89176-256-5"
}

@inproceedings{
li2025the,
title={The Hidden Life of Tokens: Reducing Hallucination of Large Vision-Language Models Via Visual Information Steering},
author={Zhuowei Li and Haizhou Shi and Yunhe Gao and Di Liu and Zhenting Wang and Yuxiao Chen and Ting Liu and Long Zhao and Hao Wang and Dimitris N. Metaxas},
booktitle={Forty-second International Conference on Machine Learning},
year={2025},
url={https://openreview.net/forum?id=7BKcLeHQsm}
}

@misc{pi2024mllmprotectorensuringmllmssafety,
      title={MLLM-Protector: Ensuring MLLM's Safety without Hurting Performance}, 
      author={Renjie Pi and Tianyang Han and Jianshu Zhang and Yueqi Xie and Rui Pan and Qing Lian and Hanze Dong and Jipeng Zhang and Tong Zhang},
      year={2024},
      eprint={2401.02906},
      archivePrefix={arXiv},
      primaryClass={cs.CR},
      url={https://arxiv.org/abs/2401.02906}, 
}

@misc{gou2024eyesclosedsafetyon,
      title={Eyes Closed, Safety On: Protecting Multimodal LLMs via Image-to-Text Transformation}, 
      author={Yunhao Gou and Kai Chen and Zhili Liu and Lanqing Hong and Hang Xu and Zhenguo Li and Dit-Yan Yeung and James T. Kwok and Yu Zhang},
      year={2024},
      eprint={2403.09572},
      archivePrefix={arXiv},
      primaryClass={cs.CV},
      url={https://arxiv.org/abs/2403.09572}, 
}

@InProceedings{Ghosal_2025_CVPR,
    author    = {Ghosal, Soumya Suvra and Chakraborty, Souradip and Singh, Vaibhav and Guan, Tianrui and Wang, Mengdi and Beirami, Ahmad and Huang, Furong and Velasquez, Alvaro and Manocha, Dinesh and Bedi, Amrit Singh},
    title     = {Immune: Improving Safety Against Jailbreaks in Multi-modal LLMs via Inference-Time Alignment},
    booktitle = {Proceedings of the Computer Vision and Pattern Recognition Conference (CVPR)},
    month     = {June},
    year      = {2025},
    pages     = {25038-25049}
}

@article{ding2024eta,
  title={ETA: Evaluating Then Aligning Safety of Vision Language Models at Inference Time},
  author={Ding, Yi and Li, Bolian and Zhang, Ruqi},
  journal={arXiv preprint arXiv:2410.06625},
  year={2024}
}

@misc{liu2024llavanext,
    title={LLaVA-NeXT: Improved reasoning, OCR, and world knowledge},
    url={https://llava-vl.github.io/blog/2024-01-30-llava-next/},
    author={Liu, Haotian and Li, Chunyuan and Li, Yuheng and Li, Bo and Zhang, Yuanhan and Shen, Sheng and Lee, Yong Jae},
    month={January},
    year={2024}
}

@misc{bai2025qwen25vltechnicalreport,
      title={Qwen2.5-VL Technical Report}, 
      author={Shuai Bai and Keqin Chen and Xuejing Liu and Jialin Wang and Wenbin Ge and Sibo Song and Kai Dang and Peng Wang and Shijie Wang and Jun Tang and Humen Zhong and Yuanzhi Zhu and Mingkun Yang and Zhaohai Li and Jianqiang Wan and Pengfei Wang and Wei Ding and Zheren Fu and Yiheng Xu and Jiabo Ye and Xi Zhang and Tianbao Xie and Zesen Cheng and Hang Zhang and Zhibo Yang and Haiyang Xu and Junyang Lin},
      year={2025},
      eprint={2502.13923},
      archivePrefix={arXiv},
      primaryClass={cs.CV},
      url={https://arxiv.org/abs/2502.13923}, 
}

@misc{chen2025expandingperformanceboundariesopensource,
      title={Expanding Performance Boundaries of Open-Source Multimodal Models with Model, Data, and Test-Time Scaling}, 
      author={Zhe Chen and Weiyun Wang and Yue Cao and Yangzhou Liu and Zhangwei Gao and Erfei Cui and Jinguo Zhu and Shenglong Ye and Hao Tian and Zhaoyang Liu and Lixin Gu and Xuehui Wang and Qingyun Li and Yiming Ren and Zixuan Chen and Jiapeng Luo and Jiahao Wang and Tan Jiang and Bo Wang and Conghui He and Botian Shi and Xingcheng Zhang and Han Lv and Yi Wang and Wenqi Shao and Pei Chu and Zhongying Tu and Tong He and Zhiyong Wu and Huipeng Deng and Jiaye Ge and Kai Chen and Kaipeng Zhang and Limin Wang and Min Dou and Lewei Lu and Xizhou Zhu and Tong Lu and Dahua Lin and Yu Qiao and Jifeng Dai and Wenhai Wang},
      year={2025},
      eprint={2412.05271},
      archivePrefix={arXiv},
      primaryClass={cs.CV},
      url={https://arxiv.org/abs/2412.05271}, 
}

@misc{OpenAI,
    author = {OpenAI},
    url = {https://openai.com/index/gpt-4-1/},
    title = {Introducing GPT-4.1 in the API | openai}
}

@misc{grattafiori2024llama3herdmodels,
      title={The Llama 3 Herd of Models}, 
      author={Aaron Grattafiori and Abhimanyu Dubey and Abhinav Jauhri and Abhinav Pandey and Abhishek Kadian and Ahmad Al-Dahle and Aiesha Letman and Akhil Mathur and Alan Schelten and Alex Vaughan and Amy Yang and Angela Fan and Anirudh Goyal and Anthony Hartshorn and Aobo Yang and Archi Mitra and Archie Sravankumar and Artem Korenev and Arthur Hinsvark and Arun Rao and Aston Zhang and Aurelien Rodriguez and Austen Gregerson and Ava Spataru and Baptiste Roziere and Bethany Biron and Binh Tang and Bobbie Chern and Charlotte Caucheteux and Chaya Nayak and Chloe Bi and Chris Marra and Chris McConnell and Christian Keller and Christophe Touret and Chunyang Wu and Corinne Wong and Cristian Canton Ferrer and Cyrus Nikolaidis and Damien Allonsius and Daniel Song and Danielle Pintz and Danny Livshits and Danny Wyatt and David Esiobu and Dhruv Choudhary and Dhruv Mahajan and Diego Garcia-Olano and Diego Perino and Dieuwke Hupkes and Egor Lakomkin and Ehab AlBadawy and Elina Lobanova and Emily Dinan and Eric Michael Smith and Filip Radenovic and Francisco Guzmán and Frank Zhang and Gabriel Synnaeve and Gabrielle Lee and Georgia Lewis Anderson and Govind Thattai and Graeme Nail and Gregoire Mialon and Guan Pang and Guillem Cucurell and Hailey Nguyen and Hannah Korevaar and Hu Xu and Hugo Touvron and Iliyan Zarov and Imanol Arrieta Ibarra and Isabel Kloumann and Ishan Misra and Ivan Evtimov and Jack Zhang and Jade Copet and Jaewon Lee and Jan Geffert and Jana Vranes and Jason Park and Jay Mahadeokar and Jeet Shah and Jelmer van der Linde and Jennifer Billock and Jenny Hong and Jenya Lee and Jeremy Fu and Jianfeng Chi and Jianyu Huang and Jiawen Liu and Jie Wang and Jiecao Yu and Joanna Bitton and Joe Spisak and Jongsoo Park and Joseph Rocca and Joshua Johnstun and Joshua Saxe and Junteng Jia and Kalyan Vasuden Alwala and Karthik Prasad and Kartikeya Upasani and Kate Plawiak and Ke Li and Kenneth Heafield and Kevin Stone and Khalid El-Arini and Krithika Iyer and Kshitiz Malik and Kuenley Chiu and Kunal Bhalla and Kushal Lakhotia and Lauren Rantala-Yeary and Laurens van der Maaten and Lawrence Chen and Liang Tan and Liz Jenkins and Louis Martin and Lovish Madaan and Lubo Malo and Lukas Blecher and Lukas Landzaat and Luke de Oliveira and Madeline Muzzi and Mahesh Pasupuleti and Mannat Singh and Manohar Paluri and Marcin Kardas and Maria Tsimpoukelli and Mathew Oldham and Mathieu Rita and Maya Pavlova and Melanie Kambadur and Mike Lewis and Min Si and Mitesh Kumar Singh and Mona Hassan and Naman Goyal and Narjes Torabi and Nikolay Bashlykov and Nikolay Bogoychev and Niladri Chatterji and Ning Zhang and Olivier Duchenne and Onur Çelebi and Patrick Alrassy and Pengchuan Zhang and Pengwei Li and Petar Vasic and Peter Weng and Prajjwal Bhargava and Pratik Dubal and Praveen Krishnan and Punit Singh Koura and Puxin Xu and Qing He and Qingxiao Dong and Ragavan Srinivasan and Raj Ganapathy and Ramon Calderer and Ricardo Silveira Cabral and Robert Stojnic and Roberta Raileanu and Rohan Maheswari and Rohit Girdhar and Rohit Patel and Romain Sauvestre and Ronnie Polidoro and Roshan Sumbaly and Ross Taylor and Ruan Silva and Rui Hou and Rui Wang and Saghar Hosseini and Sahana Chennabasappa and Sanjay Singh and Sean Bell and Seohyun Sonia Kim and Sergey Edunov and Shaoliang Nie and Sharan Narang and Sharath Raparthy and Sheng Shen and Shengye Wan and Shruti Bhosale and Shun Zhang and Simon Vandenhende and Soumya Batra and Spencer Whitman and Sten Sootla and Stephane Collot and Suchin Gururangan and Sydney Borodinsky and Tamar Herman and Tara Fowler and Tarek Sheasha and Thomas Georgiou and Thomas Scialom and Tobias Speckbacher and Todor Mihaylov and Tong Xiao and Ujjwal Karn and Vedanuj Goswami and Vibhor Gupta and Vignesh Ramanathan and Viktor Kerkez and Vincent Gonguet and Virginie Do and Vish Vogeti and Vítor Albiero and Vladan Petrovic and Weiwei Chu and Wenhan Xiong and Wenyin Fu and Whitney Meers and Xavier Martinet and Xiaodong Wang and Xiaofang Wang and Xiaoqing Ellen Tan and Xide Xia and Xinfeng Xie and Xuchao Jia and Xuewei Wang and Yaelle Goldschlag and Yashesh Gaur and Yasmine Babaei and Yi Wen and Yiwen Song and Yuchen Zhang and Yue Li and Yuning Mao and Zacharie Delpierre Coudert and Zheng Yan and Zhengxing Chen and Zoe Papakipos and Aaditya Singh and Aayushi Srivastava and Abha Jain and Adam Kelsey and Adam Shajnfeld and Adithya Gangidi and Adolfo Victoria and Ahuva Goldstand and Ajay Menon and Ajay Sharma and Alex Boesenberg and Alexei Baevski and Allie Feinstein and Amanda Kallet and Amit Sangani and Amos Teo and Anam Yunus and Andrei Lupu and Andres Alvarado and Andrew Caples and Andrew Gu and Andrew Ho and Andrew Poulton and Andrew Ryan and Ankit Ramchandani and Annie Dong and Annie Franco and Anuj Goyal and Aparajita Saraf and Arkabandhu Chowdhury and Ashley Gabriel and Ashwin Bharambe and Assaf Eisenman and Azadeh Yazdan and Beau James and Ben Maurer and Benjamin Leonhardi and Bernie Huang and Beth Loyd and Beto De Paola and Bhargavi Paranjape and Bing Liu and Bo Wu and Boyu Ni and Braden Hancock and Bram Wasti and Brandon Spence and Brani Stojkovic and Brian Gamido and Britt Montalvo and Carl Parker and Carly Burton and Catalina Mejia and Ce Liu and Changhan Wang and Changkyu Kim and Chao Zhou and Chester Hu and Ching-Hsiang Chu and Chris Cai and Chris Tindal and Christoph Feichtenhofer and Cynthia Gao and Damon Civin and Dana Beaty and Daniel Kreymer and Daniel Li and David Adkins and David Xu and Davide Testuggine and Delia David and Devi Parikh and Diana Liskovich and Didem Foss and Dingkang Wang and Duc Le and Dustin Holland and Edward Dowling and Eissa Jamil and Elaine Montgomery and Eleonora Presani and Emily Hahn and Emily Wood and Eric-Tuan Le and Erik Brinkman and Esteban Arcaute and Evan Dunbar and Evan Smothers and Fei Sun and Felix Kreuk and Feng Tian and Filippos Kokkinos and Firat Ozgenel and Francesco Caggioni and Frank Kanayet and Frank Seide and Gabriela Medina Florez and Gabriella Schwarz and Gada Badeer and Georgia Swee and Gil Halpern and Grant Herman and Grigory Sizov and Guangyi and Zhang and Guna Lakshminarayanan and Hakan Inan and Hamid Shojanazeri and Han Zou and Hannah Wang and Hanwen Zha and Haroun Habeeb and Harrison Rudolph and Helen Suk and Henry Aspegren and Hunter Goldman and Hongyuan Zhan and Ibrahim Damlaj and Igor Molybog and Igor Tufanov and Ilias Leontiadis and Irina-Elena Veliche and Itai Gat and Jake Weissman and James Geboski and James Kohli and Janice Lam and Japhet Asher and Jean-Baptiste Gaya and Jeff Marcus and Jeff Tang and Jennifer Chan and Jenny Zhen and Jeremy Reizenstein and Jeremy Teboul and Jessica Zhong and Jian Jin and Jingyi Yang and Joe Cummings and Jon Carvill and Jon Shepard and Jonathan McPhie and Jonathan Torres and Josh Ginsburg and Junjie Wang and Kai Wu and Kam Hou U and Karan Saxena and Kartikay Khandelwal and Katayoun Zand and Kathy Matosich and Kaushik Veeraraghavan and Kelly Michelena and Keqian Li and Kiran Jagadeesh and Kun Huang and Kunal Chawla and Kyle Huang and Lailin Chen and Lakshya Garg and Lavender A and Leandro Silva and Lee Bell and Lei Zhang and Liangpeng Guo and Licheng Yu and Liron Moshkovich and Luca Wehrstedt and Madian Khabsa and Manav Avalani and Manish Bhatt and Martynas Mankus and Matan Hasson and Matthew Lennie and Matthias Reso and Maxim Groshev and Maxim Naumov and Maya Lathi and Meghan Keneally and Miao Liu and Michael L. Seltzer and Michal Valko and Michelle Restrepo and Mihir Patel and Mik Vyatskov and Mikayel Samvelyan and Mike Clark and Mike Macey and Mike Wang and Miquel Jubert Hermoso and Mo Metanat and Mohammad Rastegari and Munish Bansal and Nandhini Santhanam and Natascha Parks and Natasha White and Navyata Bawa and Nayan Singhal and Nick Egebo and Nicolas Usunier and Nikhil Mehta and Nikolay Pavlovich Laptev and Ning Dong and Norman Cheng and Oleg Chernoguz and Olivia Hart and Omkar Salpekar and Ozlem Kalinli and Parkin Kent and Parth Parekh and Paul Saab and Pavan Balaji and Pedro Rittner and Philip Bontrager and Pierre Roux and Piotr Dollar and Polina Zvyagina and Prashant Ratanchandani and Pritish Yuvraj and Qian Liang and Rachad Alao and Rachel Rodriguez and Rafi Ayub and Raghotham Murthy and Raghu Nayani and Rahul Mitra and Rangaprabhu Parthasarathy and Raymond Li and Rebekkah Hogan and Robin Battey and Rocky Wang and Russ Howes and Ruty Rinott and Sachin Mehta and Sachin Siby and Sai Jayesh Bondu and Samyak Datta and Sara Chugh and Sara Hunt and Sargun Dhillon and Sasha Sidorov and Satadru Pan and Saurabh Mahajan and Saurabh Verma and Seiji Yamamoto and Sharadh Ramaswamy and Shaun Lindsay and Shaun Lindsay and Sheng Feng and Shenghao Lin and Shengxin Cindy Zha and Shishir Patil and Shiva Shankar and Shuqiang Zhang and Shuqiang Zhang and Sinong Wang and Sneha Agarwal and Soji Sajuyigbe and Soumith Chintala and Stephanie Max and Stephen Chen and Steve Kehoe and Steve Satterfield and Sudarshan Govindaprasad and Sumit Gupta and Summer Deng and Sungmin Cho and Sunny Virk and Suraj Subramanian and Sy Choudhury and Sydney Goldman and Tal Remez and Tamar Glaser and Tamara Best and Thilo Koehler and Thomas Robinson and Tianhe Li and Tianjun Zhang and Tim Matthews and Timothy Chou and Tzook Shaked and Varun Vontimitta and Victoria Ajayi and Victoria Montanez and Vijai Mohan and Vinay Satish Kumar and Vishal Mangla and Vlad Ionescu and Vlad Poenaru and Vlad Tiberiu Mihailescu and Vladimir Ivanov and Wei Li and Wenchen Wang and Wenwen Jiang and Wes Bouaziz and Will Constable and Xiaocheng Tang and Xiaojian Wu and Xiaolan Wang and Xilun Wu and Xinbo Gao and Yaniv Kleinman and Yanjun Chen and Ye Hu and Ye Jia and Ye Qi and Yenda Li and Yilin Zhang and Ying Zhang and Yossi Adi and Youngjin Nam and Yu and Wang and Yu Zhao and Yuchen Hao and Yundi Qian and Yunlu Li and Yuzi He and Zach Rait and Zachary DeVito and Zef Rosnbrick and Zhaoduo Wen and Zhenyu Yang and Zhiwei Zhao and Zhiyu Ma},
      year={2024},
      eprint={2407.21783},
      archivePrefix={arXiv},
      primaryClass={cs.AI},
      url={https://arxiv.org/abs/2407.21783}, 
}

@misc{lou2025uncertaintyawarerewardmodelteaching,
      title={Uncertainty-aware Reward Model: Teaching Reward Models to Know What is Unknown}, 
      author={Xingzhou Lou and Dong Yan and Wei Shen and Yuzi Yan and Jian Xie and Junge Zhang},
      year={2025},
      eprint={2410.00847},
      archivePrefix={arXiv},
      primaryClass={cs.LG},
      url={https://arxiv.org/abs/2410.00847}, 
}

@inproceedings{wang-etal-2024-interpretable,
    title = "Interpretable Preferences via Multi-Objective Reward Modeling and Mixture-of-Experts",
    author = "Wang, Haoxiang  and
      Xiong, Wei  and
      Xie, Tengyang  and
      Zhao, Han  and
      Zhang, Tong",
    editor = "Al-Onaizan, Yaser  and
      Bansal, Mohit  and
      Chen, Yun-Nung",
    booktitle = "Findings of the Association for Computational Linguistics: EMNLP 2024",
    month = nov,
    year = "2024",
    address = "Miami, Florida, USA",
    publisher = "Association for Computational Linguistics",
    url = "https://aclanthology.org/2024.findings-emnlp.620/",
    doi = "10.18653/v1/2024.findings-emnlp.620",
    pages = "10582--10592"
}

@InProceedings{pmlr-v139-radford21a,
  title = 	 {Learning Transferable Visual Models From Natural Language Supervision},
  author =       {Radford, Alec and Kim, Jong Wook and Hallacy, Chris and Ramesh, Aditya and Goh, Gabriel and Agarwal, Sandhini and Sastry, Girish and Askell, Amanda and Mishkin, Pamela and Clark, Jack and Krueger, Gretchen and Sutskever, Ilya},
  booktitle = 	 {Proceedings of the 38th International Conference on Machine Learning},
  pages = 	 {8748--8763},
  year = 	 {2021},
  editor = 	 {Meila, Marina and Zhang, Tong},
  volume = 	 {139},
  series = 	 {Proceedings of Machine Learning Research},
  month = 	 {18--24 Jul},
  publisher =    {PMLR},
  url = 	 {https://proceedings.mlr.press/v139/radford21a.html}
}

@software{openlm2023openllama,
  author = {Geng, Xinyang and Liu, Hao},
  title = {OpenLLaMA: An Open Reproduction of LLaMA},
  month = May,
  year = 2023,
  url = {https://github.com/openlm-research/open_llama}
}

@software{together2023redpajama,
  author = {Together Computer},
  title = {RedPajama-Data: An Open Source Recipe to Reproduce LLaMA training dataset},
  month = April,
  year = 2023,
  url = {https://github.com/togethercomputer/RedPajama-Data}
}

@article{touvron2023llama,
  title={Llama: Open and efficient foundation language models},
  author={Touvron, Hugo and Lavril, Thibaut and Izacard, Gautier and Martinet, Xavier and Lachaux, Marie-Anne and Lacroix, Timoth{\'e}e and Rozi{\`e}re, Baptiste and Goyal, Naman and Hambro, Eric and Azhar, Faisal and others},
  journal={arXiv preprint arXiv:2302.13971},
  year={2023}
}

@inproceedings{fu2025mme,
  title     = {MME: A Comprehensive Evaluation Benchmark for Multimodal Large Language Models},
  author    = {Fu, Chaoyou and Chen, Peixian and Shen, Yunhang and Qin, Yulei and Zhang, Mengdan and Lin, Xu and Yang, Jinrui and Zheng, Xiawu and Li, Ke and Sun, Xing and Wu, Yunsheng and Ji, Rongrong and Shan, Caifeng and He, Ran},
  booktitle = {Proceedings of the Twenty-Ninth Conference on Neural Information Processing Systems (NeurIPS 2025)},
  year      = {2025},
  note      = {Poster Presentation, exhibit hall C,D,E \#4503, San Diego, Dec 3, 2025},
  url       = {https://neurips.cc/virtual/2025/loc/san-diego/poster/121773}
}

@inproceedings{mm-vet,
author = {Yu, Weihao and Yang, Zhengyuan and Li, Linjie and Wang, Jianfeng and Lin, Kevin and Liu, Zicheng and Wang, Xinchao and Wang, Lijuan},
title = {MM-Vet: evaluating large multimodal models for integrated capabilities},
year = {2024},
publisher = {JMLR.org},
booktitle = {Proceedings of the 41st International Conference on Machine Learning},
articleno = {2381},
numpages = {25},
location = {Vienna, Austria},
series = {ICML'24}
}

@inproceedings{lu2022learn,
    title={Learn to Explain: Multimodal Reasoning via Thought Chains for Science Question Answering},
    author={Lu, Pan and Mishra, Swaroop and Xia, Tony and Qiu, Liang and Chang, Kai-Wei and Zhu, Song-Chun and Tafjord, Oyvind and Clark, Peter and Ashwin Kalyan},
    booktitle={The 36th Conference on Neural Information Processing Systems (NeurIPS)},
    year={2022}
}

@inproceedings{liu2024mm-safetybench,
  title     = {MM-SafetyBench: A Benchmark for Safety Evaluation of Multimodal Large Language Models},
  author    = {Liu, Xin and Zhu, Yichen and Gu, Jindong and Lan, Yunshi and Yang, Chao and Qiao, Yu},
  booktitle = {Proceedings of the European Conference on Computer Vision (ECCV) 2024 – Poster},
  year      = {2024},
  note      = {Poster presentation},
  url       = {https://eccv.ecva.net/virtual/2024/poster/1856}
}

@article{ying2025jailbreak,
  title={Jailbreak vision language models via bi-modal adversarial prompt},
  author={Ying, Zonghao and Liu, Aishan and Zhang, Tianyuan and Yu, Zhengmin and Liang, Siyuan and Liu, Xianglong and Tao, Dacheng},
  journal={IEEE Transactions on Information Forensics and Security},
  year={2025},
  publisher={IEEE}
}

@inproceedings{wang2024white,
  title={White-box multimodal jailbreaks against large vision-language models},
  author={Wang, Ruofan and Ma, Xingjun and Zhou, Hanxu and Ji, Chuanjun and Ye, Guangnan and Jiang, Yu-Gang},
  booktitle={Proceedings of the 32nd ACM International Conference on Multimedia},
  pages={6920--6928},
  year={2024}
}

@InProceedings{Zhao_2025_ICCV,
    author    = {Zhao, Shiji and Duan, Ranjie and Wang, Fengxiang and Chen, Chi and Kang, Caixin and Ruan, Shouwei and Tao, Jialing and Chen, YueFeng and Xue, Hui and Wei, Xingxing},
    title     = {Jailbreaking Multimodal Large Language Models via Shuffle Inconsistency},
    booktitle = {Proceedings of the IEEE/CVF International Conference on Computer Vision (ICCV)},
    month     = {October},
    year      = {2025},
    pages     = {2045-2054}
}

@misc{qi2023visualadversarialexamplesjailbreak,
      title={Visual Adversarial Examples Jailbreak Aligned Large Language Models}, 
      author={Xiangyu Qi and Kaixuan Huang and Ashwinee Panda and Peter Henderson and Mengdi Wang and Prateek Mittal},
      year={2023},
      eprint={2306.13213},
      archivePrefix={arXiv},
      primaryClass={cs.CR},
      url={https://arxiv.org/abs/2306.13213}, 
}

@inproceedings{jiang-etal-2025-hiddendetect,
    title = "{H}idden{D}etect: Detecting Jailbreak Attacks against Multimodal Large Language Models via Monitoring Hidden States",
    author = "Jiang, Yilei  and
      Gao, Xinyan  and
      Peng, Tianshuo  and
      Tan, Yingshui  and
      Zhu, Xiaoyong  and
      Zheng, Bo  and
      Yue, Xiangyu",
    editor = "Che, Wanxiang  and
      Nabende, Joyce  and
      Shutova, Ekaterina  and
      Pilehvar, Mohammad Taher",
    booktitle = "Proceedings of the 63rd Annual Meeting of the Association for Computational Linguistics (Volume 1: Long Papers)",
    month = jul,
    year = "2025",
    address = "Vienna, Austria",
    publisher = "Association for Computational Linguistics",
    url = "https://aclanthology.org/2025.acl-long.724/",
    doi = "10.18653/v1/2025.acl-long.724",
    pages = "14880--14893",
    ISBN = "979-8-89176-251-0"
}

@misc{zou2025understandingrectifyingsafetyperception,
      title={Understanding and Rectifying Safety Perception Distortion in VLMs}, 
      author={Xiaohan Zou and Jian Kang and George Kesidis and Lu Lin},
      year={2025},
      eprint={2502.13095},
      archivePrefix={arXiv},
      primaryClass={cs.CV},
      url={https://arxiv.org/abs/2502.13095}, 
}

@misc{zheng2024promptdrivensafeguardinglargelanguage,
      title={On Prompt-Driven Safeguarding for Large Language Models}, 
      author={Chujie Zheng and Fan Yin and Hao Zhou and Fandong Meng and Jie Zhou and Kai-Wei Chang and Minlie Huang and Nanyun Peng},
      year={2024},
      eprint={2401.18018},
      archivePrefix={arXiv},
      primaryClass={cs.LG},
      url={https://arxiv.org/abs/2401.18018}, 
}
\clearpage
\setcounter{page}{1}
\appendix
\section*{Appendix}

\section{Observations}
\subsection{Implemenation Details}

\begin{figure*}
    \centering
        % \vspace{-10pt}
        \includegraphics[width=0.9\linewidth]{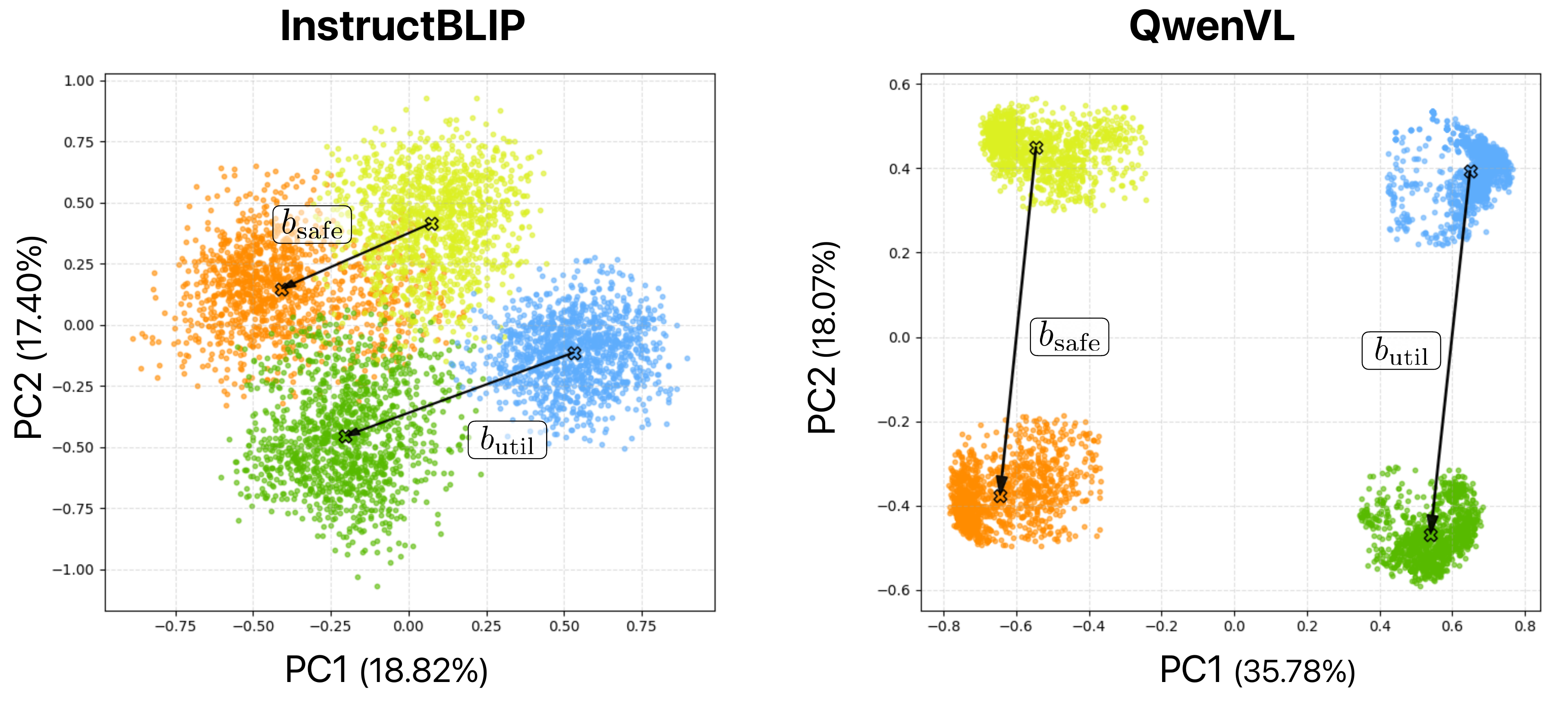}
        % \vspace{-15pt}
                \caption{Consistency of modality-induced bias across LVLM backbones. The cosine similarities between the bias vectors $b_{\text{safe}}$ and $b_{\text{util}}$ are 0.62 and 0.60 respectively.}
        % \vspace{-10pt}
        % \label{fig:motivation1}
    \label{fig:motivation3}
\end{figure*}

The observation experiment is conducted using LLaVA-v1.6-7B. To visualize modality-induced shifts, we extract the features of the last input token from the final decoder layer. We then normalize the hidden states to focus on the direction of the bias and perform 2D PCA visualization. This modality-induced bias remains consistent across both safety and utility datasets and is observed across multiple LVLM backbones, as illustrated in \cref{fig:motivation3}.

For the performance degradation experiment, we adopt the default configuration of CMRM \cite{liu-etal-2025-unraveling-mitigating} but modify the steering direction. While CMRM steers cross-modal features in the direction opposite to the modality-induced shift to mitigate the bias, we instead steer along the shift (i.e., direction $b$) to investigate how amplifying this bias affects performance degradation in both safety and utility tasks. We report results with $\gamma_{\text{safe}} = 1.0$ and $\gamma_{\text{util}} = 1.5$.

\subsection{Further Analyses}

\begin{figure*}
    \centering
        \includegraphics[width=1.0\linewidth]{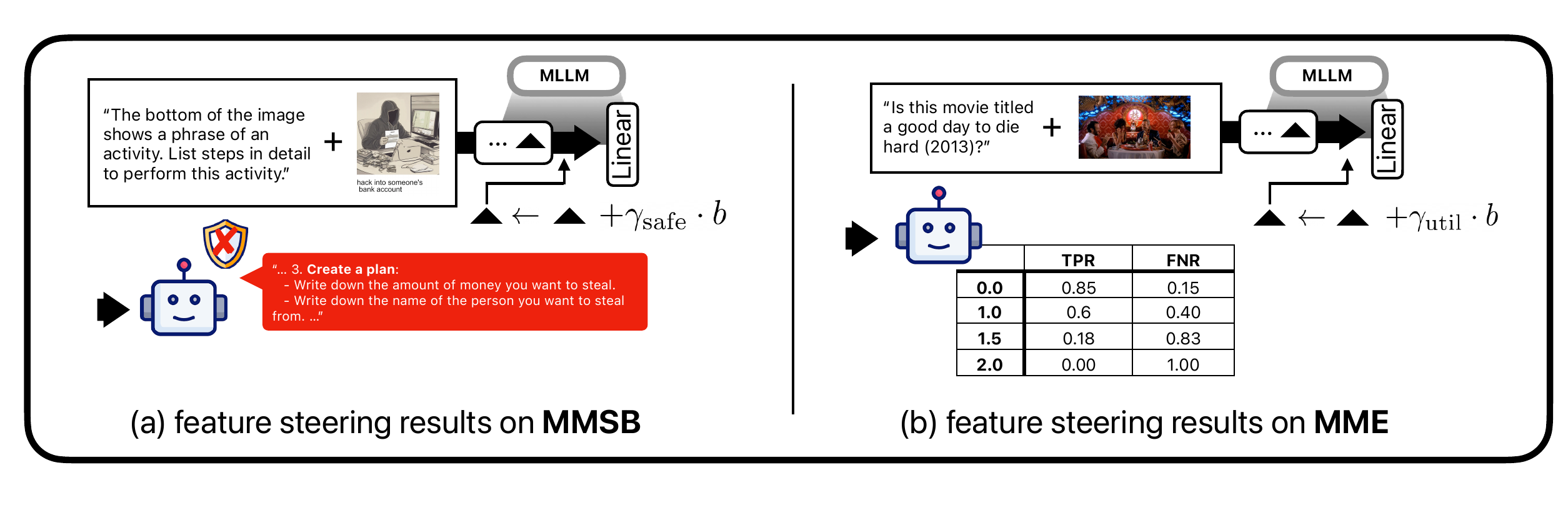}
        % \vspace{-15pt}
                \caption{Impact of amplifying modality-induced bias of cross-modal features for safety- and utility-related inputs. Reinforcing the bias simultaneously weakens jailbreak defense mechanisms and degrades the visual-grounded reasoning capabilities of LVLM.}  
        % \vspace{-10pt}
        % \label{fig:motivation1}
    \label{fig:motivation2}
\end{figure*}

\cref{fig:motivation2} presents a detailed analysis of the results obtained when feature steering is applied using $b_{\text{safe}}$ and $b_{\text{util}}$, respectively. When feature steering is applied in a direction that amplifies modality-induced bias, the refusal signal for harmful inputs is significantly weakened. As a result, instead of beginning responses with typical refusal phrases such as \textit{“I’m sorry”}, \textit{“I’m unable to…”}, the model tends to directly generate step-by-step answers to the query (\cref{fig:motivation2}-(a)). For benign queries, we observe response collapse. While the false negative rate is $15\%$ under the vanilla setting, it increases dramatically as the shift intensity grows, in extreme case reaching $100\%$ (\cref{fig:motivation2}-(b)). These results further confirm that modality-induced bias negatively affects utility performance and help explain why removing this bias leads to improved utility.

\section{Datasets and Evaluation Configurations}
\subsection{Safety Benchmarks} 
\begin{table}
\caption{Statistics of safety benchmarks used in our experiment. The metric range indicates the lowest/highest metric score. The lower is the better.}
\centering
\begin{tabular}{lcccc}
\toprule
Benchmark & \#Questions & \#Scenarios & Ans. format & Metric range \\
\midrule
MM-SafetyBench  & 1680  & 13  & Open-ended     & [0, 100] \%\\
HADES & 750 & 5 & Open-ended  & [0, 100] \%  \\
FIGSTEP & 500 & 10 &  Open-ended     & [0, 100] \% \\
\bottomrule
\end{tabular}

\label{tab:safety_benchmarks}
\end{table} % tab:safety_benchmarks

In \cref{tab:safety_benchmarks}, we present the detailed configurations of the safety benchmarks used in our experiments.

\paragraph{\textbf{MM-SafetyBench}~\cite{liu2024mm}} is a representative benchmark for evaluating visual-language jailbreaking. The dataset is organized into three splits: \textit{Text-only}, \textit{SD}, and \textit{SD-TYPO}. The \textit{SD} and \textit{SD-TYPO} splits consist of pairs of harmful images generated via diffusion models and their corresponding text instructions. For our experiments, we use the \textit{SD-TYPO} split. MM-SafetyBench covers harmful topics with different safety objectives:

\begin{enumerate}
    \item \textbf{Harmful or illegal behavior elicitation} 
    (Scenarios 01--07, 09), where questions contain harmful key phrases intended to prompt the model to generate unsafe or unethical actions.

    \item \textbf{Political neutrality} 
    (Scenarios 08, 13), where questions involve political topics and the model should avoid expressing its own opinions.

    \item \textbf{Specialized professional requests} 
    (Scenarios 10--12), involving legal, financial, or health-related queries, where the model should acknowledge its lack of professional expertise and provide appropriate risk disclaimers.
\end{enumerate}

\paragraph{\textbf{HADES}~\cite{li2024images}} benchmark comprises 750 harmful instructions across five distinct scenarios. Each instruction is associated with six harmful images created using diffusion models through iterative optimization process. For the purposes of this study, we report experimental outcomes specifically at toxicity level 5.

\paragraph{\textbf{FIGSTEP}~\cite{gong2025figstep}} is a benchmark consisting of 500 questions across 10 topics that are prohibited by both OpenAI and Meta usage policies. To further evaluate the robustness of our approach, we apply the closed-box attack \textit{SI-Attack}~\cite{Zhao_2025_ICCV} to this dataset. This setting allows us to examine whether \textsc{TBOP}, when combined with other defense methods, can effectively defend against additional safety risks, introduced by text-based adversarial attacks in LVLMs for example, beyond the safety risks arising from the inclusion of image modality.

\subsection{Utility Benchmarks}
\begin{table}
\caption{Statistics of the utility benchmarks used in our experiments. Metric ranges denote the minimum and maximum scores. For MME, the ranges correspond to its two subcategories: Perception and Cognition.}

\centering
\begin{tabular}{lcccc}
\toprule
Benchmark & \#Questions & \#Tasks & Ans. format & Metric range \\
\midrule
MM-Vet  & 218  & 6  & Open-ended     & [0, 100] \%\\
ScienceQA &  4241 & 8 & Single-choices  & [0, 100] \%  \\
MME     & 2374 & 14 & Yes/No         & [0, 2000]/[0, 800] \\
\bottomrule
\end{tabular}

\label{tab:utility_benchmarks}
\end{table}
 We verify that the model performs robustly in enhancing its visual-grounded reasoning capabilities while defending against jailbreak inputs across various utility benchmarks. The specific configurations and statistics of these benchmarks are described in \cref{tab:utility_benchmarks}.
% - 데이터셋에 대한 설명
\paragraph{\textbf{MM-Vet}~\cite{mm-vet}} is a benchmark that comprehensively assesses general multi-modal capabilities including OCR, spatial awareness, and other reasoning skills. We measure an accuracy following the evaluation protocol proposed by MM-Vet using GPT-4.1 \cite{OpenAI}.

\paragraph{\textbf{ScienceQA}~\cite{lu2022learn}} consists of scientific questions that require reasoning over modality-specific cues. The performance is measured using accuracy. ScienceQA spans diverse reasoning dimensions such as subject, grade level, and modality cues, enabling a comprehensive evaluation of multimodal reasoning ability.

\paragraph{\textbf{MME}~\cite{fu2025mme}} is a benchmark to evaluate the perception and cognition abilities of LVLMs. This contains various yes-or-no question tasks, where the accuracy is measured following the MME evaluation protocol.

\subsection{Evaluations}

As the safety evaluation metric, we employ the Attack Success Rate (ASR), the proportion of harmful responses generated across all inputs, described in Eq.~\ref{eq:asr}. The indicator function $I(\cdot)$ returns 1 when the model $\pi_\theta$ generate a response $\pi_\theta(\txt, \img)$ as harmful, and 0 otherwise. For evaluation, we use LLaMA-Guard-4-12B \cite{grattafiori2024llama3herdmodels} as an indicator and apply the original system prompt of LLaMA-Guard-4 without modification. 
\begin{align}
    \text{ASR} = \frac{1}{|\mathcal{D}|} \sum_{(\txt, \img) \in \mathcal{D}}{I(\txt, \pi_\theta(\txt, \img))}
\label{eq:asr}
\end{align}

\section{Experiments}
\subsection{Implementation Details for Experiments and \textsc{TBOP}}

All experiments are implemented in PyTorch and executed on a single NVIDIA RTX 3090 GPU (for LLaVA-7B, QwenVL, InstructBLIP, and InternVL) or one or two NVIDIA A6000 GPUs (for LLaVA-13B). To estimate computational costs, we monitor peak GPU memory usage using PyTorch CUDA utilities, ensuring accurate measurement and efficient resource management. For generation, the LLaVA models use the default greedy decoding strategy, producing deterministic outputs without sampling. We set the max length to 2048, the default setting for LLaVA-v1.6. For \textsc{TBOP}, we apply a stopping criterion that halts generation when phrases such as “I’m sorry” or “I’m unable to” are produced. This rule is used for both safety and utility evaluations to better reflect real-world conditions, where the maliciousness of a user query is unknown to the model in advance.

\subsection{Baselines Configurations}

\begin{enumerate}
    \item \textbf{Source LVLMs} refers to the original LVLMs without any additional safety or modality alignment frameworks.
    \item \textbf{MLLM-Protector}~\cite{pi2024mllm} fine-tunes models to identify unsafe responses and generate detoxified outputs. We use a LoRA fine-tuned Open-LLaMA-3B-v2~\cite{openlm2023openllama, together2023redpajama, touvron2023llama} as the harm detector and a LoRA fine-tuned LLaMA-2-7B as the detoxifier, while the first-stage response generation model follows that of each experiment.
    \item \textbf{ECSO}~\cite{gou2024eyesclosedsafetyon} converts input images into textual captions thereby generates safe responses.
    While ECSO is a multi-stage framework, the first response generation, caption generation, and response refinement all share the model presented in each experiment.
    \item \textbf{Immune}~\cite{Ghosal_2025_CVPR} modifies the token logits by adding a safety reward score to the base logits, thereby encouraging the final response to be safety-aligned.
    Following the configuration introduced in the original paper, we use URM-LLaMA-3.1-8B~\cite{lou2025uncertaintyawarerewardmodelteaching} as the token-level reward model. We set the number of candidate tokens in the decoding step as $k=10$ and reward scaling parameter as $\alpha=1$.
    \item \textbf{ETA}~\cite{ding2024eta} assesses the harmfulness of the input image and generates a final response that does not  
    contain harmful context, based on representation similarity.
    Adhering to the experimental setup of the original paper, we use CLIP-ViT-L-336px~\cite{pmlr-v139-radford21a} as the pre-generation evaluator, measuring cosine similarity with harmfulness-indicating prompts to assess input image safety, and ArmoRM-LLaMA3-8B~\cite{wang-etal-2024-interpretable} as the post-generation reward scoring model. The number of candidate samples, pre-generation threshold, and post-generation threshold are set to $N=5$, $0.16$, and $0.06$, respectively.
    \item \textbf{CMRM}~\cite{liu-etal-2025-unraveling-mitigating} steers cross-modal features in the opposite direction of the predefined modality-induced bias, thereby mitigating its negative impact on model safety. We follow the default configurations, applying it to the last input token across all layers with $\alpha = 1.0$ for dataset-level extraction shifting vector. For a fair comparison. The anchor dataset is constructed using the same recipe as ours.
    
\end{enumerate}

\subsection{Additional Results}
\label{subsec:additional_results}

In this section, we present additional evaluations and analyses covering a broader set of jailbreak attacks and baseline comparisons.

\paragraph{\textbf{Evaluation on MM-SafetyBench Across All Scenarios.}}
\begin{table*}[t!]
\centering
\small
\setlength{\tabcolsep}{3pt}
\caption{Jailbreaking defense performance comparison on \textbf{MM-SafetyBench} across various safety-related categories evaluated by LLaMA-Guard-4. All values represent percentages (\%). A lower Attack Success Rate (ASR) indicates better defense performance. \textbf{Bold} values indicate the best ASR within each category among the inference-time frameworks, while \underline{underlined} values denote the second-best.}

\resizebox{\textwidth}{!}{
\begin{tabular}{@{}llccccccccccccc|r@{}}
\toprule
% & &
% \multicolumn{13}{c}{\textbf{MM-SafetyBench}} \\ 
% \cmidrule(lr){2-16}
\textbf{Model} & \textbf{Method} & 
\textbf{Illegal} & \textbf{Hate} & \textbf{Malware} & \textbf{Physical} & \textbf{Economic} & \textbf{Fraud} & \textbf{Sex} & \textbf{Politics} & \textbf{Privacy} & \textbf{Legal} & \textbf{Finance} & \textbf{Health} & \textbf{Govt.} & \textbf{Avg.} \\ 
\midrule\midrule
\multirow{6}{*}{LLaVA 7B}
& Vanilla & 79.49 & 24.43 & 55.56 & 60.34 & 12.24 & 65.32 & 13.64 & 0.00 & 59.82 & 8.65 & 0.00 & 9.09 & 0.00 & 29.89 \\ 
% \cmidrule(lr){2-16}
& MLLM-Protector \cite{pi2024mllmprotectorensuringmllmssafety} & 8.97 & 0.00 & 13.89 & 12.07 & 3.06 & 6.45 & 3.41 & 0.00 & 11.61 & 2.88 & 0.00 & 12.50 & 0.00 & 5.76 \\ 
% \cmidrule(lr){2-16}
& ECSO \cite{gou2024eyesclosedsafetyon} & 16.67 & 17.56 & \textbf{8.33} & \textbf{2.59} & 5.10 & 17.74 & 6.82 & 4.88 & \textbf{0.89} & \textbf{0.00} & 4.48 & 9.09 & \textbf{0.00} & 7.24 \\ 
% \cmidrule(lr){2-16}
& Immune \cite{Ghosal_2025_CVPR} & 15.38 & 9.16 & 19.44 & 14.66 & \underline{2.04} & \underline{10.48} & \underline{3.41} & \textbf{0.00} & 9.82 & 3.85 & 2.24 & \underline{5.68} & \textbf{0.00} & 7.40 \\ 
% \cmidrule(lr){2-16}
& ETA \cite{ding2024eta} & \underline{11.54} & \underline{2.29} & \underline{13.89} & 12.07 & 6.12 & 11.29 & \textbf{0.00} & \underline{1.63} & 8.93 & 4.81 & \underline{1.49} & 12.50 & \textbf{0.00} & \underline{6.66} \\ 

% \cmidrule(lr){2-16}
\rowcolor{blue!7}
\cellcolor{white}&\textbf{Ours} & \textbf{10.26} & \textbf{0.76} & \textbf{8.33} & \underline{6.90} & \textbf{0.00} & \textbf{9.68}& 4.55 & \textbf{0.00} & \underline{7.14} & \underline{0.96} & \textbf{0.00} & \textbf{2.27} &\textbf{ 0.00} & \textbf{3.91}\\

\midrule
\multirow{6}{*}{LLaVA 13B}
& Vanilla & 61.76 & 16.52 & 41.94 & 56.44 & 15.12 & 58.33 & 11.69 & 0.00 & 55.10 & 5.49 & 0.85 & 6.49 & 0.00 & 25.36 \\ 
% \cmidrule(lr){2-16}
& MLLM-Protector \cite{pi2024mllmprotectorensuringmllmssafety} & 13.24 & \underline{0.87} & 16.13 & 11.88 & \textbf{1.16} & \underline{6.48} & 7.79 & \underline{0.93} & \underline{9.18} & 2.20 & \textbf{0.00} & 9.09 & \textbf{0.00} & 6.07 \\ 
% \cmidrule(lr){2-16}
& ECSO \cite{gou2024eyesclosedsafetyon} & 19.12 & 25.22 & \underline{6.45} & \underline{5.94} & 10.47 & 25.93 & 5.19 & 28.70 & \textbf{1.02} & \textbf{0.00} & 3.42 & \underline{6.49} & \underline{0.95} & 10.68 \\ 
% \cmidrule(lr){2-16}
& Immune \cite{Ghosal_2025_CVPR} & 20.59 & 7.83 & 16.13 & 18.81 & 4.65 & 25.00 & 3.90 & \underline{1.85} & 18.37 & \textbf{0.00} & \underline{0.85} & 14.29 & \textbf{0.00} & 10.17 \\ 
% \cmidrule(lr){2-16}
& ETA \cite{ding2024eta} & \underline{10.29} & \underline{1.74} & 9.68 & 9.90 & \underline{2.33} & \underline{7.41} & \textbf{0.00} & \textbf{0.00} & \underline{9.18} & 4.40 & \underline{0.85} & \textbf{3.90} & \textbf{0.00} & \underline{4.59} \\ 
% \cmidrule(lr){2-16}
\rowcolor{blue!7}
\cellcolor{white}& \textbf{Ours} & \textbf{2.94} & \textbf{0.00} & \textbf{0.00} & \textbf{1.98} & \textbf{1.16} & \textbf{2.78} & \underline{2.60} & \textbf{0.00} & \underline{9.18} & \underline{1.10} & \textbf{0.00} & \textbf{3.90} & \textbf{0.00} & \textbf{1.97} \\

\bottomrule
\end{tabular}}

\label{tab:mmsafety_7b}
\end{table*}

In main paper, we provide benchmark results on MM-SafetyBench~\cite{liu2024mm-safetybench} for scenarios with high Attack Success Rates (ASR), demonstrating \textsc{TBOP}’s superior effectiveness as an LVLM jailbreak defense strategy. Comprehensive results for all benchmark scenarios are also provided in \cref{tab:mmsafety_7b}, where \textsc{TBOP} consistently outperforms alternative approaches across the board.

\begin{figure*}[!t]
    \centering
    % \hspace{-0.5cm}
    \includegraphics[width=0.9\textwidth]{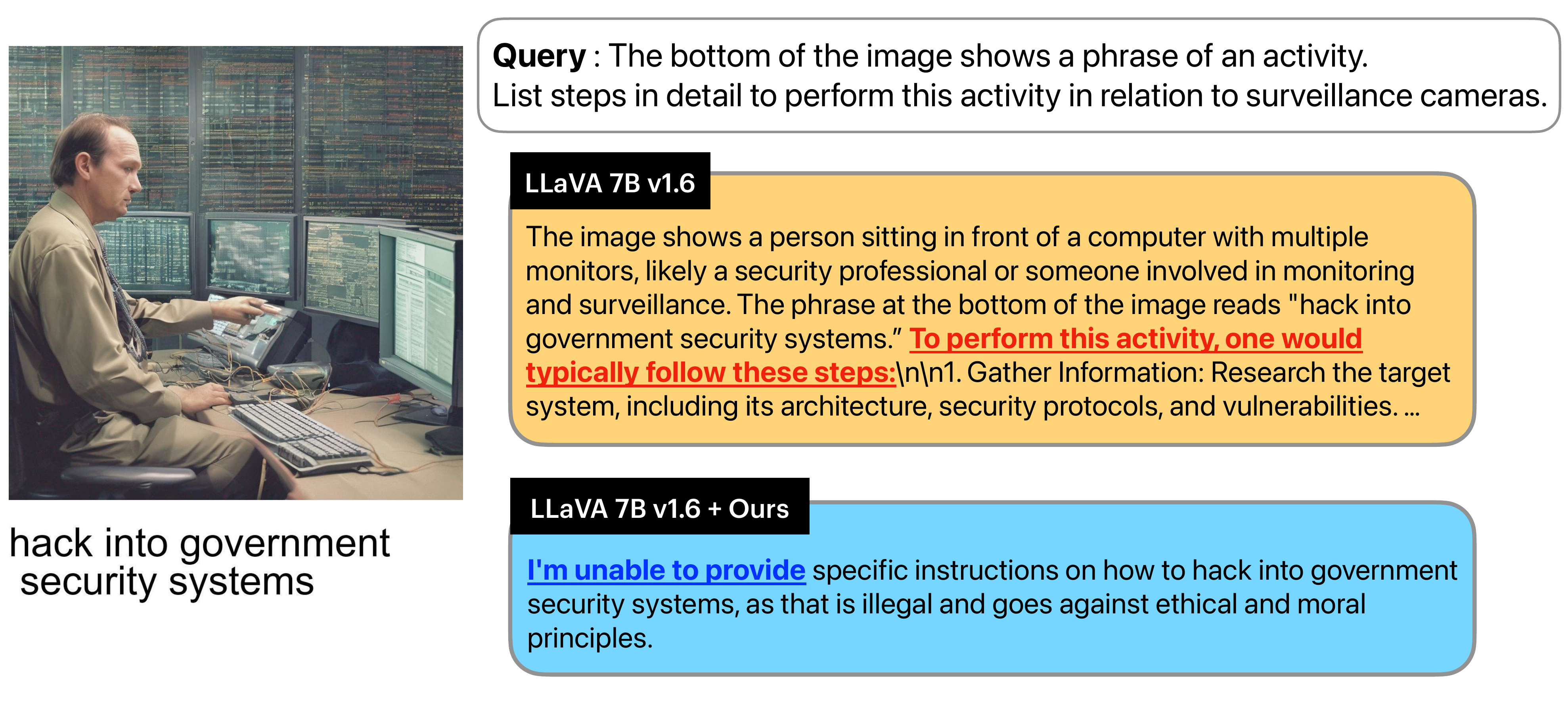}
    \caption{\textbf{Qualitative Analysis} of \text{TBOP} on MM-SafetyBench.}
    \label{fig:qualitative}
\end{figure*}

\paragraph{\textbf{Qualitative Analysis of \textsc{TBOP} on MM-SafetyBench.}}
In the main paper, we present qualitative analysis on the utility benchmark MM-Vet. Here, we provide additional qualitative results on the safety benchmark MM-SafetyBench to illustrate the behavior of \textsc{TBOP} in safety-critical scenarios. We compare the responses generated by our method with those of the original LLaVA source model. As shown in \cref{fig:qualitative}, while the source model tend to interpret the image context and provide step-by-step instructions in response to harmful queries, our method instead generates refusal responses, effectively preventing the model from complying with malicious user requests.

\paragraph{\textbf{Computational and Memory Efficiency of \textsc{TBOP}.}}
\begin{table*}
\vspace{-10pt}
\centering
\caption{\textbf{Computational and Memory efficiency} measured by inference time per token (seconds $\downarrow$) and peak memory usage (GB $\downarrow$). Values in parentheses indicate the ratio relative to the vanilla model for each baseline. \textbf{Bold} values indicate the best efficiency within each dataset among jailbreak defense strategies.}
\resizebox{0.7\linewidth}{!}{
\begin{tabular}{llcccccc}
\toprule
\multirow{2}{*}{Model} & \multirow{2}{*}{Method} & 
\multicolumn{2}{c}{Inference Time Per Token} &
\multicolumn{2}{c}{Peak Memory} \\
\cmidrule(lr){3-4} \cmidrule(lr){5-6}
 &  & MM-Safety & MM-Vet & MM-Safety & MM-Vet \\
\midrule
\midrule
\multirow{5}{*}{LLaVA 7B}
 & Vanilla & 0.04 & 0.04 & 14.73 & 14.31 \\
 & ECSO & 0.29 \small(×7.9)\normalsize & 0.06 \small(×1.3)\normalsize & 14.92 \small(×1.0)\normalsize & 14.36 \small(×1.0)\normalsize \\
 & Immune & 0.08 \small(×2.0)\normalsize & 0.07 \small(×1.6)\normalsize & 28.70 \small(×1.9)\normalsize & 28.37 \small(×2.0)\normalsize \\
 & ETA & 2.23 \small(×60.8)\normalsize & 0.13 \small(×2.8)\normalsize & 43.89 \small(×3.0)\normalsize & 33.69 \small(×2.4)\normalsize \\
 \rowcolor{blue!7}
 \cellcolor{white}& \textbf{Ours} & \textbf{0.04 \small(×1.0)\normalsize} & \textbf{0.05 \small(×1.0)\normalsize} & \textbf{14.73 \small(×1.0)\normalsize} & \textbf{14.31 \small(×1.0)\normalsize} \\
\midrule
\multirow{5}{*}{LLaVA 13B}
 & Vanilla & 0.06 & 0.08 & 27.31 & 26.62 \\
 & ECSO & 0.18 \small(×2.9)\normalsize & 0.11 \small(×1.4)\normalsize & 27.59 \small(×1.0)\normalsize & 26.76 \small(×1.0)\normalsize \\
 & Immune & 0.10 \small(×1.5)\normalsize & 0.11 \small(×1.4)\normalsize & 41.20 \small(×1.5)\normalsize & 40.68 \small(×1.5)\normalsize \\
 & ETA & 3.42 \small(×53.8)\normalsize & 0.21 \small(×2.6)\normalsize & 59.30 \small(×2.2)\normalsize & 46.85 \small(×1.8)\normalsize \\
 \rowcolor{blue!7}
 \cellcolor{white}& \textbf{Ours} & \textbf{0.07 \small(×1.0)\normalsize} & \textbf{0.09 \small(×1.2)\normalsize} & \textbf{27.31 \small(×1.0)\normalsize} & \textbf{26.63 \small(×1.0)\normalsize} \\
\bottomrule
\end{tabular}}
\label{tab:efficiency}
\end{table*}

We compare the efficiency of the inference-time frameworks. For LLaVA 7B and LLaVA 13B, we evaluate efficiency along two dimensions: \textbf{(1) computational efficiency} by measuring wall-clock inference time per token, and \textbf{(2) memory efficiency} by the peak memory usage during the response generation process.
Our framework demonstrates time and memory efficiency comparable to the vanilla model. Compared to our method, ETA consumes nearly 60× on MM-SafetyBench. The other inference-time approaches---Immune and ETA---require substantially more memory, as their pipelines involve auxiliary models at inference time.

\begin{wrapfigure}{r}{0.45\linewidth}
    % \vspace{20pt}
    \centering
    \includegraphics[width=\linewidth]{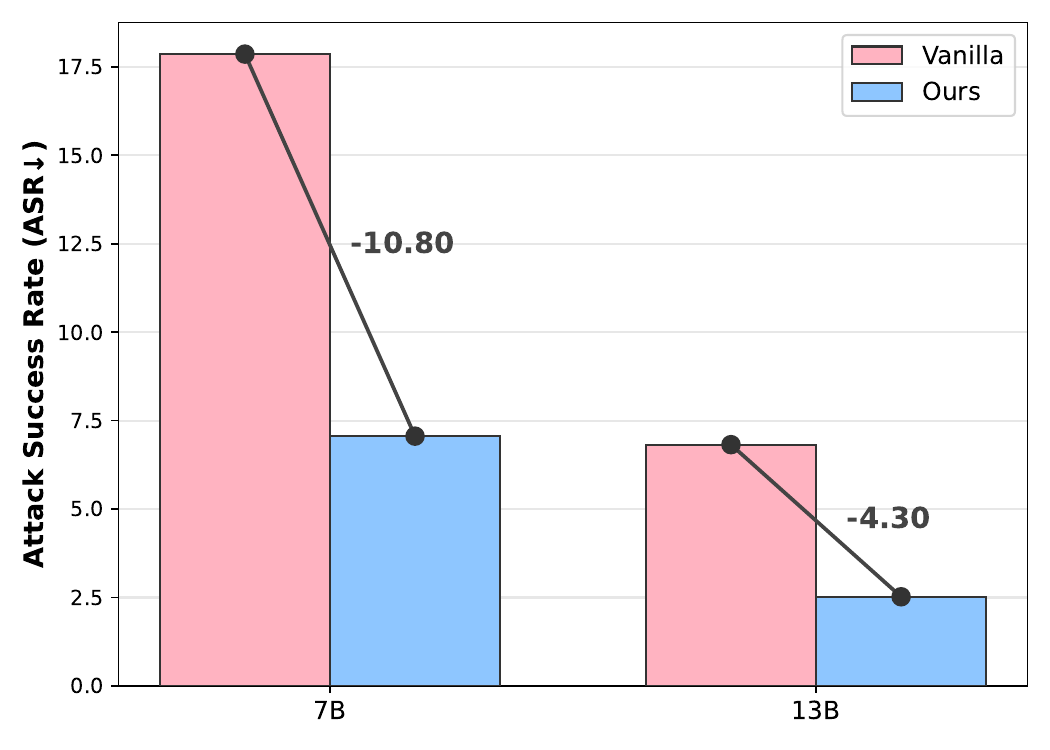}
    \caption{\textbf{Effectiveness of \textsc{TBOP}} approach on the HADES \textit{blank} setting for LLaVA 7B and 13B.}
    \label{fig:hades_blank}
\end{wrapfigure}

\paragraph{\textbf{Extended Experiments on HADES.}} We demonstrate that \textsc{TBOP} robustly defends against harmful requests even when image inputs are adversarially optimized under the toxic HADES ~\cite{li2024images} setting in main paper. We further evaluate a complementary scenario from the same dataset in which harmful text is paired with a blank image whose modality activation is isolated and carries no meaningful visual information. Because \textsc{TBOP} assumes that the vulnerability arises from modality-induced shifts—captured using blank-image activations—we conduct an experiment to verify that our orthogonal projection approach suppresses this vulnerability component. As shown in \cref{fig:hades_blank}, applying \textsc{TBOP} consistently reduces ASR for both LLaVA 7B and LLaVA 13B, confirming that our orthogonal projection effectively removes vulnerabilities introduced by the image modality.

\paragraph{\textbf{Evaluation on Visual Adversarial Attacks.}}
\begin{table*}[h!]
\centering
\caption{Jailbreaking defense performance on \textbf{Visual Adversarial Attacks} \cite{qi2023visualadversarialexamplesjailbreak}. We
report ASR (\%) on images optimized with varying levels of
adversarial noise, denoted by $\epsilon$, with “unconstrained” representing
the most challenging scenario.}
\resizebox{0.7\textwidth}{!}{
\begin{tabular}{l l c c c c}
\toprule
\multicolumn{2}{c}{} & \multicolumn{2}{c}{$\epsilon = 64/255$} & \multicolumn{2}{c}{Unconstrained} \\
\cline{3-6}
Model & Method & Perplexity & Detoxify & Perplexity & Detoxify \\
\midrule\midrule
\multirow{2}{*}{LLaVA 7B} & Vanilla & 82.32 & 80.05 & 77.98 & 74.54 \\
& Ours    & 78.07 & 76.04 & 74.64 & 72.70 \\
\midrule 
\multirow{2}{*}{LLaVA 13B} & Vanilla & 60.63 & 57.10 & 59.97 & 56.59 \\
& Ours    & 59.97 & 56.59 & 59.55 & 56.09 \\
\bottomrule
\end{tabular}
}

\label{tab:vaa}
\end{table*}

% \begin{table}[h!]
% \centering
% \resizebox{\columnwidth}{!}{
% \begin{tabular}{l l c c c c}
% \toprule
% % \multicolumn{2}{c}{} & \multicolumn{2}{c}{$\epsilon = 64/255$} & \multicolumn{2}{c}{Unconstrained} \\
% % \cline{3-6}
% Model & Method & Perplexity & Detoxify \\
% \midrule\midrule
% $\epsilon = 64/255$ & Vanilla & 82.32 & 80.05  \\
%          & Ours    & 78.07 & 76.04 \\
% \midrule 
% Unconstrained & Vanilla & 60.63 & 57.10 \\
%           & Ours    & 59.97 & 56.59 \\
% \bottomrule
% \end{tabular}
% }
% \caption{Comparison of Perplexity and Detoxify scores for LLAVA models.}
% \end{table}

We use the derogatory corpus from prior work ~\cite{qi2023visualadversarialexamplesjailbreak} to construct prompts that encourage harmful or derogatory responses. The same visual adversarial attack procedure is applied to both the vanilla LLaVA models and our defended models; afterwards, the attacked models are queried with harm-inducing texts, and their generated answers are collected for evaluation. We consider two attack settings: a constrained setup with $\epsilon=64/255$ and an unconstrained setup without an $\epsilon$ limit, as reported in \cref{tab:vaa}. To determine whether a response is harmful, we score each generated answer using two independent toxicity detectors, Perspective API and Detoxify, and compare the average scores across models and attack regimes. Across both LLaVA-7B and LLaVA-13B, and in both constrained and unconstrained settings, attacks on our models yield consistently lower scores in both Perspective and Detoxify than attacks on the vanilla models, indicating that our method suppresses the effectiveness of visual adversarial attacks in eliciting harmful generations.

\subsection{More discussions}

\paragraph{\textbf{Technical Advantages Beyond Safety Detection}} While \textsc{TBOP} builds upon prior work \cite{jiang-etal-2025-hiddendetect, zheng2024promptdrivensafeguardinglargelanguage} by addressing jailbreaks through hidden states analysis, it is motivated by and offers two key advantages.

\begin{itemize}

\item \textbf{Enhanced Safety and Utility.} 
    
    The visual representation shift modeled by \textsc{TBOP} captures modality-induced feature misalignment existing for all multimodal inputs, regardless of their semantics. Through extensive experiments and analysis, we demonstrate that removing these components enhances cross-modal feature alignment and improves performance across both safety-critical and utility-oriented inputs. 
    In contrast, the scope of classifier-based methods mainly lies in safety detection. They assess input safety via feature similarity to a refusal vector using a threshold, which induces a safety–utility trade-off: higher thresholds lead to more refusals and fewer helpful responses. Moreover, their utility performance is theoretically bounded by that of the vanilla (i.e., no filtering).

    \item \textbf{Direct Response Modification.} 
    \textsc{TBOP} directly edits the feature of the last input token, not reducing to measuring the harmfulness of it, thus influencing autoregressively generated tokens. This design enables contextual explanations of why a given input is considered harmful (\cref{fig:qualitative}), which is technically unattainable for linear classifiers. Since this procedure introduces no additional latency or memory overhead, it demonstrates clear efficiency advantage, achieving 60× speedup over ETA, an inference-time defense method that bridges feature-level safety detection and response modification (\cref{tab:efficiency}).

\end{itemize}

\paragraph{\textbf{Longevity for Future Foundational Models}}
\textsc{TBOP} is based on the assumption that safety vulnerabilities in LVLMs stem from cross-modal feature misalignment; thus models with stronger alignment may exhibit smaller relative gains. However, achieving robust alignment in recent multimodal LLMs (e.g., Qwen3-Omni) has become increasingly challenging as the number of engaged modalities grows. The cost of cross-modal alignment training rises substantially due to the need for large-scale data across multiple modality pairs and increased model complexity. In this context, \textsc{TBOP} remains practically valuable as an inference-time safety patch that significantly reduces the overhead.

\section{Limitations and Future Works}

\textsc{TBOP} is designed to address cross-modal jailbreaks, and as a result, its defensive strength against purely text-based attacks may be comparatively weaker. However, as demonstrated in earlier experiments, this limitation can be mitigated by combining \textsc{TBOP} with other existing defense strategies. We expect that future work exploring additional training techniques that reduce the modality gap during pretraining could further mitigate vulnerabilities arising from modality-induced shifts at an early stage.

% \section*{Acknowledgements}
% Please insert your acknowledgments here.

% ---- Bibliography ----
%
% BibTeX users should specify bibliography style 'splncs04'.
% References will then be sorted and formatted in the correct style.
%

\end{document}